\documentclass[letterpaper]{article} 
\usepackage{aaai25}  
\usepackage{times}  
\usepackage{helvet}  
\usepackage{courier}  
\usepackage[hyphens]{url}  
\usepackage{graphicx} 
\usepackage{natbib}  
\usepackage{caption} 
\makeatletter
\def\blfootnote{\gdef\@thefnmark{}\@footnotetext}
\makeatother

\title{\ourmethodns: Causal Discovery from Non-Stationary Time Series}
\author{ 
    Sarah Mameche\textsuperscript{\rm 1},
    Lénaïg Cornanguer\textsuperscript{\rm 1}, 
    Urmi Ninad\textsuperscript{\rm 2\rm 3}\equalcontrib,
    Jilles Vreeken\textsuperscript{\rm 1}\equalcontrib
}
\affiliations{
    \textsuperscript{\rm 1}CISPA Helmholtz Center for Information Security\\
    \textsuperscript{\rm 2}Technische Universität Berlin, Germany\\
    \textsuperscript{\rm 3}German Aerospace Center, Institute of Data Science\\
    sarah.mameche@cispa.de, 
    lenaig.cornanguer@cispa.de,
    urmi.ninad@tu-berlin.de,
    jv@cispa.de
}

\makeatletter
\newif\if@restonecol
\makeatother

\usepackage{ntheorem}
\usepackage{latexsym}
\usepackage{subcaption}
\usepackage{graphicx,centernot}
\usepackage[ruled, vlined, nofillcomment, linesnumbered, algo2e]{algorithm2e}
\usepackage{algorithm}
\usepackage{algorithmic}

\usepackage[notheorems]{eda} 
\usepackage{prettyplots}
\usepackage{footmisc}

\usepackage{amsfonts}
\usepackage{amssymb}

\usepackage{mathtools}

\usepackage{thmtools}
\usepackage{thm-restate}
\usepgfplotslibrary{fillbetween}
\usepackage{pgfplotstable}
\usepackage{pgfplots}
\usepgfplotslibrary{colormaps}%
\usetikzlibrary{pgfplots.statistics}
\usetikzlibrary{pgfplots.colormaps}
\usetikzlibrary{shapes.geometric}
\usetikzlibrary{shapes.multipart}
\pgfplotsset{
        compat=1.14,
    }
\usetikzlibrary{calc}
\usetikzlibrary{patterns}
\usetikzlibrary{positioning}  
\usetikzlibrary{pgfplots.groupplots}
\let\oldcap\cap
\let\oldcup\cup
\let\oldemptyset\emptyset
\usepackage{mathabx}
\let\cap\oldcap
\let\cup\oldcup
\let\emptyset\oldemptyset
\usepackage{scalerel}

\usepackage{prettyplots} 
\graphicspath{ {./figs/} }
\ifpdf
\usetikzlibrary{external}
\fi

\newcommand{\ourmethodns}{\textsc{SpaceTime}}
\newcommand{\ourmethod}{\textsc{SpaceTime~}}
\newcommand{\ourmethodeos}{\textsc{SpaceTime}} 

\newcommand{\indep}{\perp \!\!\! \perp}

\newcommand{\norm}[1]{\left\lVert#1\right\rVert}

\newcommand{\T}{\mathcal T}
\newcommand{\D}{\mathcal D} 
\newcommand{\W}{s}  
\newcommand{\C}{\mathcal C}
\newcommand{\Ci}[1]{C_#1} 
\newcommand{\R}{\mathcal R}
\newcommand{\Ri}[1]{R_#1}
\newcommand{\cps}{\mathcal L} 
\newcommand{\ri}{r}

\newcommand{\ci}{k}
\newcommand{\cj}{k'}
\newcommand{\cN}{K}
\newcommand{\Wi}{\W} 
\newcommand{\Wj}{\W'} 
\newcommand{\ix}{i}
\newcommand{\xj}{j}
\newcommand{\xN}{m}
\newcommand{\ti}{t}
\newcommand{\tN}{n}

\newcommand{\XX}{\mathbf X} 
\newcommand{\XT}{\big( \XX_t \big)_{t \in \T}} 
 
\newcommand{\XTD}{\big( \big( \XX_t^d \big)_{t \in \T}\big)_{d \in \D}} 
\newcommand{\XTDs}{  \XX_\T^\D } 
\newcommand{\XTs}{  \XX_\T } 
\newcommand{\Xt}[2]{X_{(#1) (#2)}}

\newcommand{\iX}[1]{X_{(#1)}}
\newcommand{\Xtc}[3]{X_{#2}^{#3}}

\newcommand{\Gg}{\mathcal G}
 
\newcommand{\G}{\mathcal{G}_{\taumax}}
\newcommand{\Gwindow}{\mathcal{G}_{\taumax}} 
\newcommand{\FCM}{\mathcal M} 
 
\newcommand{\Gs}{\mathcal{G}^{\ast}_{\taumax}}

\DeclareMathOperator{\PA}{\textit{pa}}

\newcommand{\Ntc}[3]{N_{#2}^{#3}}
\newcommand{\taumax}{\boldsymbol \tau }
\newcommand{\fcr}[3]{f^{(#2, #3)}} 
 
\newcommand{\ficr}[3]{f_{(#1)}^{(#2, #3)}} 
 
\newcommand{\thetacr}[2]{\theta^{(#1, #2)}}

\newcommand{\nback}[1][-.95pt]{
  \mathrel{\raisebox{#1}{$\rotatebox[origin=c]{-315}{\scaleobj{0.55}{-}}$}}
}
\newcommand{\undernegpreccurlyeq}{%
\mathrel{\ooalign{$\preccurlyeq$\cr\kern1.2pt$\nback$}}}

\newcommand{\undernegsuccurlyeq}{%
\mathrel{\ooalign{$\succcurlyeq$\cr\kern1.2pt$\nback$}}}

\newcommand{\M}{\mathcal H} 
\newcommand{\kernel}{\kappa}
\newcommand{\Mk}{\M_\kernel}

\newcommand{\dynotears}{\textsc{DYNOTEARS}~}
\newcommand{\cdnod}{\textsc{CD\nobreakdash-NOD}~}
\newcommand{\varlingam}{\textsc{VARLiNGAM}~}
\newcommand{\pcmcip}{\textsc{PCMCI+}~}
\newcommand{\rpcmci}{\textsc{R\nobreakdash-PCMCI}~}
\newcommand{\jcmcip}{\textsc{J\nobreakdash-PCMCI+}~}
\newcommand{\dynotearsns}{\textsc{DYNOTEARS}}
\newcommand{\cdnodns}{\textsc{CD\nobreakdash-NOD}}
\newcommand{\varlingamns}{\textsc{VARLiNGAM}}
\newcommand{\pcmcipns}{\textsc{PCMCI+}}

\newcommand{\jcmcipns}{\textsc{J\nobreakdash-PCMCI+}}

\definecolor{tsne-vio}{RGB}{131,24,76}
\tikzset{
	invis_node/.style={draw=white!0}
}
    \tikzset{
        dag_edge/.style={  ->,>=stealth, black, line width=1, },
        w1/.style={ ->,>=stealth, black!10, line width=.8, },
        w2/.style={   ->,>=stealth, black!20, line width=.8,  },
        w3/.style={ ->,>=stealth, black!30, line width=.8,  },
        w4/.style={ ->,>=stealth, black!50, line width=.8, }, 
        strongest/.style={ ->,>=stealth, line width=.8, black
        } 
    }

\pgfplotsset{ 
    colormap={ourmap}{
        rgb255=(251,191,36) 
        rgb255=(22,161,74) 
        rgb255=(3,105,161) 
        rgb255=(131,24,76)
    },
} 
\pgfplotsset{ 
    colormap={ourmap2}{
        rgb255=(241, 193, 75), 
        rgb255=(162, 178, 75), 
        rgb255=(80, 162, 83), 
        rgb255=(38, 86, 128), 
        rgb255=(79, 46, 82), 
        rgb255=(185, 18, 61), 
        rgb255=(174, 42, 63), 
    },
} 

\newsavebox{\tempbox}
\newcommand{\textbox}[1]
{\savebox{\tempbox}{#1}
 \ifdim\wd\tempbox<4cm\relax
   \makebox[4cm]{\usebox{\tempbox}}%
 \else
   \parbox{4cm}{\raggedright #1}%
 \fi}

\pgfdeclarelayer{background}
\pgfdeclarelayer{foreground}
\pgfsetlayers{background,main,foreground}

\definecolor{color(mmci)}{cmyk}{1,0.4,0,0}
\definecolor{color(cispa)}{rgb}{0.008,0.356,0.584}
\definecolor{color(helmholtzblue)}{RGB}{0, 90, 160}
\definecolor{color(helmholtzgreen)}{RGB}{140, 180, 35}

\definecolor{wong-blue}{RGB}{0, 114, 178}
\definecolor{wong-orange}{RGB}{230, 159, 0}
\definecolor{wong-green}{RGB}{0, 158, 115}
\definecolor{wong-reddishpurple}{RGB}{204, 121, 167}
\definecolor{wong-skyblue}{RGB}{86, 180, 233}
\definecolor{wong-vermillion}{RGB}{213, 94, 0}
\definecolor{wong-yellow}{RGB}{240, 228, 66}

\definecolor{tol-q-bright-blue}{HTML}{4477AA}
\definecolor{tol-q-bright-cyan}{HTML}{66CCEE}
\definecolor{tol-q-bright-green}{HTML}{228833}
\definecolor{tol-q-bright-yellow}{HTML}{CCBB44}
\definecolor{tol-q-bright-red}{HTML}{EE6677}
\definecolor{tol-q-bright-purple}{HTML}{AA3377}
\definecolor{tol-q-bright-grey}{HTML}{BBBBBB}

\definecolor{tol-q-high-contrast-white}{HTML}{FFFFFF}
\definecolor{tol-q-high-contrast-yellow}{HTML}{DDAA33}
\definecolor{tol-q-high-contrast-red}{HTML}{BB5566}
\definecolor{tol-q-high-contrast-blue}{HTML}{004488}
\definecolor{tol-q-high-contrast-black}{HTML}{000000}

\definecolor{tol-q-vibrant-blue}{HTML}{0077BB}
\definecolor{tol-q-vibrant-cyan}{HTML}{33BBEE}
\definecolor{tol-q-vibrant-teal}{HTML}{009988}
\definecolor{tol-q-vibrant-orange}{HTML}{EE7733}
\definecolor{tol-q-vibrant-red}{HTML}{CC3311}
\definecolor{tol-q-vibrant-magenta}{HTML}{EE3377}
\definecolor{tol-q-vibrant-grey}{HTML}{BBBBBB}

\definecolor{tol-q-muted-indigo}{HTML}{332288}
\definecolor{tol-q-muted-cyan}{HTML}{88CCEE}
\definecolor{tol-q-muted-teal}{HTML}{44AA99}
\definecolor{tol-q-muted-green}{HTML}{117733 }
\definecolor{tol-q-muted-olive}{HTML}{999933}
\definecolor{tol-q-muted-sand}{HTML}{DDCC77}
\definecolor{tol-q-muted-rose}{HTML}{CC6677}
\definecolor{tol-q-muted-wine}{HTML}{882255}
\definecolor{tol-q-muted-purple}{HTML}{AA4499}
\definecolor{tol-q-muted-pale-grey}{HTML}{DDDDDD}

\definecolor{tol-mark-pale-blue}{HTML}{BBCCEE}
\definecolor{tol-mark-pale-cyan}{HTML}{CCEEFF}
\definecolor{tol-mark-pale-green}{HTML}{CCDDAA}
\definecolor{tol-mark-pale-yellow}{HTML}{EEEEBB}
\definecolor{tol-mark-pale-red}{HTML}{FFCCCC}
\definecolor{tol-mark-pale-grey}{HTML}{DDDDDD}

\definecolor{tol-mark-dark-blue}{HTML}{222255}
\definecolor{tol-mark-dark-cyan}{HTML}{225555}
\definecolor{tol-mark-dark-green}{HTML}{225522}
\definecolor{tol-mark-dark-yellow}{HTML}{666633}
\definecolor{tol-mark-dark-red}{HTML}{663333}
\definecolor{tol-mark-pale-grey}{HTML}{555555}

\definecolor{tol-q-light-blue}{HTML}{77AADD}
\definecolor{tol-q-light-cyan}{HTML}{99DDFF}
\definecolor{tol-q-light-mint}{HTML}{44BB99}
\definecolor{tol-q-light-pear}{HTML}{BBCC33}
\definecolor{tol-q-light-olive}{HTML}{AAAA00}
\definecolor{tol-q-light-yellow}{HTML}{EEDD88}
\definecolor{tol-q-light-orange}{HTML}{EE8866}
\definecolor{tol-q-light-pink}{HTML}{FFAABB}
\definecolor{tol-q-light-pale-grey}{HTML}{DDDDDD}

\colorlet{color(ourmethod)}{tol-q-muted-indigo} 	
\colorlet{color(ourmethodsemi)}{tol-q-muted-indigo} 	
\colorlet{color(asso)}{tol-q-muted-cyan} 	
\colorlet{color(grecond)}{tol-q-muted-teal} 	
\colorlet{color(nmf)}{tol-q-muted-green} 	
\colorlet{color(nmfr)}{tol-q-muted-olive} 	
\colorlet{color(sofa)}{tol-q-muted-sand} 	
\colorlet{color(binaps)}{tol-q-muted-rose} 	
\colorlet{color(pimp)}{tol-q-muted-wine}

\definecolor{TSneColor(0.0)}{RGB}{136, 204, 238}
\definecolor{TSneColor(1.0)}{RGB}{117, 194, 214}
\definecolor{TSneColor(2.0)}{RGB}{98, 185, 190}
\definecolor{TSneColor(3.0)}{RGB}{79, 175, 166}
\definecolor{TSneColor(4.0)}{RGB}{62, 164, 140}
\definecolor{TSneColor(5.0)}{RGB}{47, 149, 112}
\definecolor{TSneColor(6.0)}{RGB}{33, 135, 83}
\definecolor{TSneColor(7.0)}{RGB}{19, 121, 54}
\definecolor{TSneColor(8.0)}{RGB}{26, 98, 72}
\definecolor{TSneColor(9.0)}{RGB}{35, 74, 96}
\definecolor{TSneColor(10.0)}{RGB}{45, 50, 120}
\definecolor{TSneColor(11.0)}{RGB}{67, 50, 134}
\definecolor{TSneColor(12.0)}{RGB}{115, 98, 130}
\definecolor{TSneColor(13.0)}{RGB}{163, 146, 125}
\definecolor{TSneColor(14.0)}{RGB}{210, 193, 120}
\definecolor{TSneColor(15.0)}{RGB}{206, 193, 104}
\definecolor{TSneColor(16.0)}{RGB}{187, 178, 85}
\definecolor{TSneColor(17.0)}{RGB}{168, 164, 66}
\definecolor{TSneColor(18.0)}{RGB}{156, 150, 55}
\definecolor{TSneColor(19.0)}{RGB}{171, 135, 74}
\definecolor{TSneColor(20.0)}{RGB}{185, 121, 94}
\definecolor{TSneColor(21.0)}{RGB}{199, 107, 113}
\definecolor{TSneColor(22.0)}{RGB}{191, 89, 113}
\definecolor{TSneColor(23.0)}{RGB}{172, 70, 103}
\definecolor{TSneColor(24.0)}{RGB}{153, 51, 94}
\definecolor{TSneColor(25.0)}{RGB}{137, 35, 87}
\definecolor{TSneColor(26.0)}{RGB}{147, 45, 106}
\definecolor{TSneColor(27.0)}{RGB}{156, 54, 125}
\definecolor{TSneColor(28.0)}{RGB}{166, 64, 144}
\definecolor{TSneColor(29.0)}{RGB}{178, 92, 164}
\definecolor{TSneColor(30.0)}{RGB}{192, 135, 183}
\definecolor{TSneColor(31.0)}{RGB}{207, 178, 202}
\definecolor{TSneColor(32.0)}{RGB}{221, 221, 221}

\definecolor{clust1col}{RGB}{101,66,134}
\definecolor{clust2col}{HTML}{0891b2}
\definecolor{nipsvio}{RGB}{101,66,134}
\definecolor{testblue}{HTML}{06b6d4}

\definecolor{s0}{HTML}{ff9b02} 
\definecolor{s1}{HTML}{F9D923} 
\definecolor{s2}{HTML}{134e6f} 
\definecolor{s3}{HTML}{ff6150} 
\definecolor{s4}{HTML}{1ac0c6} 
\definecolor{s5}{HTML}{36AE7C} 

\definecolor{e0}{HTML}{ff6150}
\definecolor{e1}{HTML}{1ac0c6}
\definecolor{e2}{HTML}{F9D923}
\definecolor{e3}{HTML}{36AE7C}
\definecolor{e4}{HTML}{134e6f}
\definecolor{e5}{HTML}{ff6150}
\definecolor{e6}{HTML}{1ac0c6}
\definecolor{e7}{HTML}{ff9b02}

\definecolor{abst1}{HTML}{E4C1F9}
\definecolor{abst2}{HTML}{4F7CAC}
\definecolor{abst3}{HTML}{9EEFE5}
\definecolor{abst4}{HTML}{541388} 

\definecolor{mygreen}{rgb}{0.0, 0.5, 0.0}
\definecolor{mygray}{rgb}{0.3, 0.3, 0.3}

\definecolor{variolila}{RGB}{115,100,137}
\definecolor{lilacgray}{RGB}{152,150,164}
\definecolor{red1}{RGB}{201,59,69}
\definecolor{cc1}{rgb}{0.59, 0.78, 0.64}
\definecolor{niceblue}{RGB}{3, 79, 132}
\definecolor{iceblue}{RGB}{128, 182, 207}
\definecolor{cc1}{rgb}{0.59, 0.78, 0.64}
\definecolor{niceblue}{RGB}{3, 79, 132}
\definecolor{iceblue}{RGB}{128, 182, 207}
\definecolor{ca3}{rgb}{0.83, 0.69, 0.22}
\definecolor{red1}{RGB}{201,59,69}
\definecolor{red2}{RGB}{169,22,48}
\definecolor{red3}{RGB}{249,129,115}
\definecolor{cl1}{rgb}{0.61, 0.77, 0.89} 
\definecolor{cl2}{rgb}{0.64, 0.68, 0.82} 
\definecolor{cl3}{rgb}{0.85, 0.65, 0.13} 
\definecolor{ca1}{rgb}{0.6, 0.81, 0.93} 
\definecolor{ca2}{RGB}{139, 204, 109} 
\definecolor{rosequartz}{RGB}{247,202,201}
\definecolor{serenity}{RGB}{145,168,209}
\definecolor{peachecho}{RGB}{247,120,107}
\definecolor{snorkelblue}{RGB}{3,79,132}
\definecolor{buttercup}{RGB}{250,224,60}
\definecolor{limpetshell}{RGB}{152,221,222}
\definecolor{lilacgray}{RGB}{152,150,164}
\definecolor{fiesta}{RGB}{221,65,50}
\definecolor{icedcoffee}{RGB}{177,143,106}
\definecolor{greenflash}{RGB}{121,199,83}
\definecolor{purple}{RGB}{153,0,153}
\definecolor{turquise}{RGB}{0,153,153}
\definecolor{poop}{RGB}{203,178,52}
\definecolor{blue1}{RGB}{215, 216, 233}
\definecolor{blue2}{RGB}{243, 243, 248}
\definecolor{varioblue}{RGB}{68, 114, 157}
\definecolor{variogreen}{RGB}{125, 171, 113}
 
\definecolor{slate50}{HTML}{f8fafc} 
\definecolor{slate100}{HTML}{f1f5f9} 
\definecolor{slate200}{HTML}{e2e8f0}
\definecolor{slate300}{HTML}{cbd5e1} 
\definecolor{slate400}{HTML}{94a3b8} 
\definecolor{slate500}{HTML}{64748b} 
\definecolor{cmap1}{RGB}{251,191,36}
\definecolor{cmap2}{RGB}{22,161,74}
\definecolor{cmap3}{RGB}{3,105,161}
\definecolor{cmap4}{RGB}{131,24,76} 

\pgfplotsset{
	/tikz/normal shift/.code 2 args = {%
		\pgftransformshift{%
			\pgfpointscale{#2}{\pgfplotspointouternormalvectorofticklabelaxis{#1}}%
		}%
	},%
	eda line/.style={
		tick align        	= outside,
		scaled ticks      	= false,
		enlargelimits     	= false,
		ticklabel shift   	= {3pt},
		axis lines*       	= left,
		line cap          	= round,
		clip              	= false,
		tick style    		= {thin, black, major tick length=2pt},
		x tick label style 	= {font=\scriptsize, yshift = 1pt},
		y tick label style 	= {font=\scriptsize, xshift = 1pt},
		xtick style       	= {normal shift={x}{3pt}},
		ytick style       	= {normal shift={y}{3pt}},
		x axis line style 	= {thick,normal shift={x}{3pt}},
		y axis line style 	= {thick,normal shift={y}{3pt}},
		x label style 		= {font=\scriptsize},
		y label style 		= {font=\scriptsize},
	}
}

\makeatletter
\renewtheoremstyle{plain}
  {\item{\theorem@headerfont ##1\ ##2\theorem@separator}~}
  {\item{\theorem@headerfont ##1\ ##2\ (##3)\theorem@separator}~}
\makeatother

\declaretheoremstyle[
spaceabove=\topsep,
spacebelow=\topsep,
headfont=\normalfont\bfseries,    
notefont=\normalfont\itshape,     
notebraces={(}{)},                
bodyfont=\normalfont,             
qed=\qedsymbol]{thmstyle}

\declaretheorem[style=thmstyle,name=Theorem]{theorem} 

{\theoremheaderfont{\upshape\bfseries}
  \theorembodyfont{\normalfont\slshape}
  \newtheorem{definition}{Definition}}
{\theoremheaderfont{\upshape\bfseries}
  \theorembodyfont{\normalfont\slshape} 
  \newtheorem{assumption}{Assumption}}
{\theoremheaderfont{\upshape\bfseries}
  \theorembodyfont{\normalfont\slshape}
  }
{\theoremheaderfont{\upshape\bfseries}
 	}
{\theoremheaderfont{\upshape\bfseries}
  \theorembodyfont{\normalfont\slshape}
  }
 {\theoremheaderfont{\upshape\bfseries}
 \theorembodyfont{\normalfont\slshape}
 }

\usepackage{tikz,amsmath} 

\usetikzlibrary{arrows,snakes,backgrounds,patterns,matrix,shapes,fit,calc,shadows,plotmarks,positioning}

\begin{document} 
\clearpage 

\maketitle

\blfootnote{\emph{Accepted at the 39th Annual AAAI Conference on Artificial Intelligence (AAAI 2025).}}
\begin{abstract}
Understanding causality is challenging and often complicated by changing causal relationships over time and across environments.  
Climate patterns, for example, shift over time with recurring seasonal trends, 
while also depending on geographical characteristics such as ecosystem variability.  
Existing methods for discovering causal graphs from time series either assume stationarity, do not permit both temporal and spatial distribution changes, or are unaware of locations with the same causal relationships. 
In this work, we therefore unify the three tasks of causal graph discovery in the non-stationary multi-context setting, of reconstructing temporal regimes, and of partitioning datasets and time intervals into those where invariant causal relationships hold. To construct a consistent score that forms the basis of our method, we employ the Minimum Description Length  principle. Our resulting algorithm  \ourmethod simultaneously accounts for heterogeneity across space and non-stationarity over time. Given multiple time series, it discovers regime changepoints and a temporal causal graph using non-parametric functional modeling and kernelized discrepancy testing. We also show that our method provides insights into real-world phenomena such as river-runoff measured at different catchments and biosphere-atmosphere interactions across ecosystems.

\end{abstract}

\section{Introduction} 
\label{sec:intro}
Gaining insight into the dynamics of complex real-world processes is closely tied to understanding the causal mechanisms underlying their function. Causal models~\cite{pearl:09:causalitybook} allow reasoning about the effect of intervention and distributional change, which makes them especially useful for modeling systems under changing environments or over time. At odds with this, however, is that established methods for discovering causal networks from data~\cite{spirtes:00:pc,chickering:02:ges} often make the simplifying assumption that all samples have a fixed data-generating process.  

\tikzset{ 
    dag_node/.style={draw=none, thick, rounded corners, circle,
        minimum size=0.4cm, fill=slate200
        }
}

\def\yshiftts{8}
\def\poscontextts{6}
\def\poslabelts{\poscontextts+18}
\def\xshiftts{\poslabelts+12}
\def\colorregimeone{dollarbill}
\def\colorregimetwo{goldenrod}
\def\colorcontextone{mambacolor4}
\def\colorcontexttwo{nipsvio}

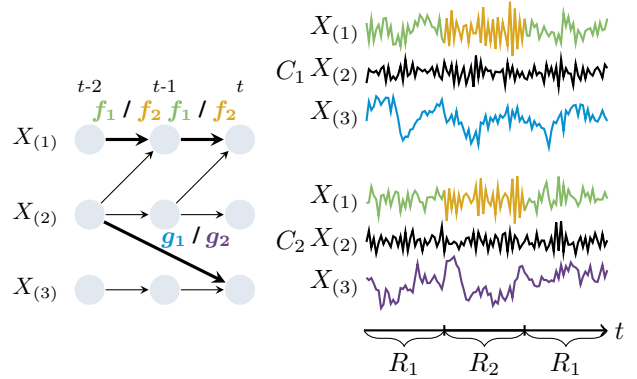
\begin{figure}[h!]
    \centering
    \begin{subfigure}[c]{0.4\linewidth}
        
        \begin{tikzpicture}[->,>=stealth, node distance=1cm, every node/.style={draw, thick, font=\small}]
        
        \node (A1) [dag_node] {};
        \node (A2) [dag_node, right of=A1] {};
        \node (A3) [dag_node, right of=A2] {};
        \node (B1) [dag_node, below of=A1] {};
        \node (B2) [dag_node, right of=B1] {};
        \node (B3) [dag_node, right of=B2] {};
        \node (C1) [dag_node, below of=B1] {};
        \node (C2) [dag_node, right of=C1] {};
        \node (C3) [dag_node, right of=C2] {};
        
        \foreach \i/\j in {1/2, 2/3} {
            \path (A\i) edge[very thick] node[midway, above, draw=none, fill=none, xshift=0pt, yshift=3pt,font=\bfseries\boldmath\small] {{\textcolor{\colorregimeone}{$f_1$} / \textcolor{\colorregimetwo}{$f_2$}}} (A\j);
            \path (B\i) edge (B\j);
            \path (C\i) edge (C\j);
        }
        
        \path (B1) edge (A2);
        \path (B2) edge (A3);
        \path (B1) edge[very thick] node[midway, above, draw=none, fill=none, xshift=12pt, 
        yshift=-2pt,font=\bfseries\boldmath\small] {{\textcolor{\colorcontextone}{$g_1$} / \textcolor{\colorcontexttwo}{$g_2$}}} (C3);
        
        \node[draw=none] at ($(A1.west) - (0.5,0)$) {$X_{(1)}$};
        \node[draw=none] at ($(B1.west) - (0.5,0)$) {$X_{(2)}$};
        \node[draw=none] at ($(C1.west) - (0.5,0)$) {$X_{(3)}$};
        
        \node[draw=none] at ($(A1.north) + (0, 0.5)$) {\scriptsize {$t$-$2$}};
        \node[draw=none] at ($(A2.north) + (0, 0.5)$) { \scriptsize {$t$-$1$}};
        \node[draw=none] at ($(A3.north) + (0, 0.5)$) {\scriptsize {$t$}};
        
        \end{tikzpicture}
    \end{subfigure}
    \hfill
    \begin{subfigure}[c]{0.575\linewidth}
        \centering
        \begin{tikzpicture}
        \begin{axis}[
            scale only axis,
            width=\linewidth,
            height=5.5cm,
            axis x line=none,  
            axis y line=none,  
            xtick=\empty,
            ytick=\empty,
            xmin=0
        ]
        
        
        
        \node[draw=none] at (axis cs:\poslabelts,0) {$X_{(1)}$}; 
        \addplot[\colorregimeone, mark=none, unbounded coords=jump, x filter/.expression={x>33 && x<66 ? nan : x+\xshiftts}] table [col sep=comma, x=t, y=a] {figs/illustrations/timeseries1.csv};
        
        \addplot[\colorregimetwo, mark=none, unbounded coords=jump, x filter/.expression={x<33 || x>66 ? nan : x+\xshiftts}] table [col sep=comma, x=t, y=a] {figs/illustrations/timeseries1.csv};
        
        \addplot[black, mark=none, y filter/.expression={y-\yshiftts}, x filter/.expression={x+\xshiftts}]  table [col sep=comma, x=t, y=b] {figs/illustrations/timeseries1.csv};
        \node[draw=none] at (axis cs:\poslabelts,-\yshiftts) {$X_{(2)}$}; 

        
        \node[draw=none] at (axis cs:\poslabelts,-2*\yshiftts) {$X_{(3)}$};
        \addplot[\colorcontextone, mark=none, y filter/.expression={y-(\yshiftts*2)}, x filter/.expression={x+\xshiftts}] table [col sep=comma, x=t, y=c] {figs/illustrations/timeseries1.csv};

        \node[draw=none] at (axis cs:\poscontextts,-\yshiftts) {$C_1$};

        
               
        \node[draw=none] at (axis cs:\poslabelts,-4*\yshiftts) {$X_{(1)}$}; 
        \addplot[\colorregimeone, mark=none, y filter/.expression={y-4*\yshiftts}, unbounded coords=jump, x filter/.expression={x>33 && x<66 ? nan : x+\xshiftts}] table [col sep=comma, x=t, y=a] {figs/illustrations/timeseries2.csv};
        
        \addplot[\colorregimetwo, mark=none, y filter/.expression={y-4*\yshiftts}, unbounded coords=jump, x filter/.expression={x<33 || x>66 ? nan : x+\xshiftts}] table [col sep=comma, x=t, y=a] {figs/illustrations/timeseries2.csv};
               
        \addplot[black, mark=none, y filter/.expression={y-5*\yshiftts}, x filter/.expression={x+\xshiftts}]  table [col sep=comma, x=t, y=b] {figs/illustrations/timeseries2.csv};
        \node[draw=none] at (axis cs:\poslabelts,-5*\yshiftts) {$X_{(2)}$};

        
        \node[draw=none] at (axis cs:\poslabelts,-6*\yshiftts) {$X_{(3)}$};
        \addplot[\colorcontexttwo, mark=none, y filter/.expression={y-(\yshiftts*6)}, x filter/.expression={x+\xshiftts}] table [col sep=comma, x=t, y=c] {figs/illustrations/timeseries2.csv};

        \node[draw=none] at (axis cs:\poscontextts,-\yshiftts*5) {$C_2$}; 

        \def\ytimeline{-7.1*\yshiftts}
        \addplot[draw=none] coordinates{(\xshiftts,\ytimeline) (\xshiftts+100,\ytimeline-3)};
        \draw[->] (axis cs:\xshiftts,\ytimeline) -- (axis cs:\xshiftts+100,\ytimeline);
        \addplot[mark=|] coordinates{(\xshiftts+33,\ytimeline) (\xshiftts+66,\ytimeline)};
        \draw[->, thin] (axis cs:\xshiftts,\ytimeline) -- (axis cs:\xshiftts+100,\ytimeline);
        \node[draw=none] at (axis cs:\xshiftts+105,\ytimeline) {$t$};
        \draw[thin, decorate,decoration={brace,amplitude=6pt,mirror}] (axis cs:\xshiftts, \ytimeline) -- (axis cs:\xshiftts+33, \ytimeline) node [black,below=5pt,midway]  {$R_1$};
        \draw[thin, decorate,decoration={brace,amplitude=6pt,mirror}] (axis cs:\xshiftts+33, \ytimeline) -- (axis cs:\xshiftts+66, \ytimeline) node [black,below=5pt,midway] {$R_2$};
        \draw[thin, decorate,decoration={brace,amplitude=6pt,mirror}] (axis cs:\xshiftts+66, \ytimeline) -- (axis cs:\xshiftts+100, \ytimeline) node [black,below=5pt,midway] {$R_1$};
        \end{axis}
        \end{tikzpicture}

    \end{subfigure}

\caption{\emph{Left}: A temporal causal graph representing the data-generating causal mechanism of three variables ($X_{(1)}$, $X_{(2)}$, and $X_{(3)}$). The bold edges indicate a local mechanism change across contexts ($g_1$ or $g_2$) or over time ($f_1$ or $f_2$). \emph{Right}: Variable measurements from different contexts ($C_1$ and $C_2$) and under different temporal regimes ($R_1$ and $R_2$).}
\label{fig:illustrations:interventions_model}
\end{figure}

This assumption is especially problematic in time series data. In climate science, for example, 
measurements often come from geographical regions  experiencing different climatic conditions. Even within a single region, weather patterns are not constant but often shift over time due to seasonality, extreme events, or global climate change. Similarly, disease trajectories of patients at multiple locations could show variability due to varying healthcare infrastructures or population heterogeneity, and change over time in response to local interventions or changes in public behavior. 

To address such scenarios, in this work we consider a collection of
non-stationary time series datasets. Fig.~\ref{fig:illustrations:interventions_model}, right shows this setting for three variables that we observe over time in two contexts. Changes in their data-generating causal mechanisms (colored) occur at unknown points in time ($\iX 1$), called \emph{changepoints}, or between contexts ($\iX 3$). 
We aim to jointly discover changes across temporal and spatial scales while explaining the underlying causal interactions, as shown in Fig.~\ref{fig:illustrations:interventions_model}, left.  

Recent methods focus either on changepoint discovery in single time series~\cite{saggioro:20:RPCMCI} or on multiple datasets without detecting changepoints~\cite{gunther:23:jpcmcip}. Here, we suggest considering both tasks simultaneously and looking for recurring, invariant causal mechanisms. 
Examples include cities in geographical regions with similar environmental conditions, or time periods with seasonal trends or economic cycles. Groups of datasets across which the causal mechanisms do not change are said to belong to the same \emph{context}, and similarly, time intervals when the causal mechanisms of all variables remain stationary are said to belong to the same \emph{regime}.
For example, in Fig.~\ref{fig:illustrations:interventions_model},
$\iX 1$ experiences a change at two points in time, but we observe a repeating regime $R_1$. 

Unlike the mostly constraint-based  causal discovery literature for non-stationary time series~\cite{runge:20:pcmcip,assad:22:ts_survey}, we rely on functional modeling to provide direct insights into causal mechanisms and their similarity. We model causal relationships non-parametrically using Gaussian processes (GPs)~\cite{gp-book:rasmussen}. 
A constraint-based approach to causal discovery is also limited by the effectiveness of conditional independence testing, which is a provably hard problem~\cite{rajen:20:hardness_citesting}. This motivates us to adopt instead a score-based approach based on sparsity constraints, where we build on the Minimum Description Length principle (MDL)~\cite{rissanen:78:mdl}. We will show how this applies to the time series causal discovery setting and changepoint detection tasks in the remainder of this work.  

\paragraph{Contributions} We propose a framework for causal modeling of non-stationary time series in heterogeneous contexts. 
We confirm theoretically that our causal discovery scoring criterion is consistent and develop the \ourmethod algorithm to efficiently discover causal models, regime changepoints, and invariances over time (regimes) and space (contexts). Besides showing that our approach works well in different synthetic data-generation settings, we study two real-world applications exploring drivers of river discharge~\cite{guenther:23:river}  and functional biosphere-atmosphere interactions~\cite{baldocchi:20:fluxdata}. Our approach discovers joint causal networks across  the span of multiple years and various locations,  
and reveals gradients in interaction strength across temporal and spatial scales that match previous findings~\cite{kraemer:20:biospherestates,krich:21:flux}.


\section{Related Work} 
 
Causal discovery in time series data has long been limited to Granger-causality~\cite{granger}, which focuses on predicting whether past values of one variable can help forecast future values of another. We focus here on related work that aims to identify true causal relationships.

Most methods, such as \pcmcipns~\cite{runge:20:pcmcip}, \varlingamns~\cite{hyvarinen:10:varlingam} as well as \dynotearsns~\cite{pamfil:20:dynotears}, assume that the data are stationary in time and space.
The problem of non-stationarity in time has been studied from different perspectives. \rpcmci~\cite{saggioro:20:RPCMCI}, like us, assumes the existence of recurrent regimes within each of which the data is stationary. \cdnodns~\cite{cdnod}, on the other hand,  can capture temporal non-stationarity if it can be modeled as a smooth function of the time index. $\text{PCMCI}_{\Omega}$~\cite{gao:23:semistationary} only considers semi-stationary time series where the causal effects occur periodically.
Finally, unlike the aforementioned approaches, \jcmcipns~\cite{gunther:23:jpcmcip} can handle multiple heterogeneous time series datasets as well as non-stationarity. However, it does not perform regime detection or context partitioning.


\section{Theoretical Framework}

In the following, we introduce the concepts of contexts and regimes and our causal modeling assumptions. 
\subsection{Problem Setting}

Throughout our work, we consider multivariate discrete-time stochastic processes over a set of continuous variables $ \XX= \{\iX 1,\ldots \iX \xN \}$. We assume that we sample each variable 
at a discrete set of time indices $\T = \{t_1, \ldots, t_\tN \}$. This results in measurements $\Xt \ix \ti $ at each time point $t$, which we denote as $X_\ti$ when the variable index can be suppressed. We denote the resulting time series as $\XT$, shorthand $\XTs$.

As in our motivating setting, we observe not just one but a collection of time series datasets $\XTD$, or $\XTDs$ for short, for a given set of dataset indices $\D$.  
To capture that certain groups of locations share  the same data-generating mechanisms while for others they differ, we propose  partitioning the index set $\D$ into different contexts.

\begin{definition}[Contexts]
	We define the contexts for a variable $X$ as partitions $\C = \{\Ci 1, \ldots, \Ci \cN\}$ of $\D$ into disjoint, nonempty index sets $\Ci \ci \neq \emptyset$, with $\cup_\ci \Ci \ci = \D$ and  $\Ci \ci  \cap C_{k'} = \emptyset $ for any $\ci \neq \cj $, 
 such that no mechanism change of $X$ occurs between datasets $d_1, d_2$ in the same $\Ci \ci$.
\end{definition} 

Analogously, to represent changes  over time, we consider a subset  $\cps \subset \T$ of points that index such changes, so-called \emph{changepoints}.
As similar trends might repeat periodically over time, there are not necessarily as many distinct causal mechanisms as changepoints. This motivates partitioning the regions defined by $\cps$ into groups, so-called regimes.  

\begin{definition}[Regimes] 
	We define the regimes for a  variable $X$ as set partitions $\R = \{ \Ri 1, \ldots, \Ri L \}$ of $\T$ into disjoint, nonempty index sets under a given set of changepoints $\cps$,
 such that $\cps=\{t \in \T \mid \emph{regime}(t) \neq \emph{regime}( t+1)\}$, where 
 no mechanism change of $X$ occurs in the same $ \Ri l$.
\end{definition}

Above, we associate contexts with datasets and regimes with time indices through the functions $\emph{context} : \D \rightarrow \C$ and $\emph{regime} : \T \rightarrow \R$, which map each dataset (resp. time index) to its corresponding context (resp. regime). Note that we have $l \in \mathcal L$ whenever the causal mechanism for \emph{at least} one node $X$ changes. By assuming a fixed set  $\cps$ over $\D$, we presume all datasets are identically affected by non-stationarity, for example, due to seasonality.

Together, the contexts and regimes determine in which subsamples of observations no mechanism change occurs. We illustrate these notions in Fig.~\ref{fig:illustrations:interventions_model}, highlighting each subsample with a different color. We write a subsample in $C_k$ and $R_r$ as $s = (k, r)$ for brevity.  

In summary, our input data is as follows.

\begin{definition}[Non-stationary Time Series]
	We consider multivariate time series $\XTDs$ with known dataset indices $d \in \mathcal{D}$ and unknown changepoints $\cps \subset \T$, as well as
  latent contexts $\C$ over $\D$ and regimes $\R$ over $\T$ for each $X$. 
\end{definition}

In the following, we will define more formally what it means that variables undergo causal mechanism changes by introducing our causal model.

\subsection{Causal Model}
As our graphical representation of causal interactions over time, we work with the following temporal causal graphs. 

\begin{definition}[Temporal Causal Graph \cite{assad:22:ts_survey}] A temporal causal graph (TCG) $\Gg =(V, E)$ over $\mathbf X$ and $\T$ is a directed acyclic graph (DAG) $\Gg$, where $V$ includes nodes $\Xt \ix t$ for each $i$ and $t$ and $E$ includes two types of edges, 
	\begin{itemize} 
		\item instantaneous links $\Xt \ix t \to \Xt \xj t$ for $i \neq j$ whenever  $X_\ix$ causes $X_\xj$ at time point $t \in \T$, and
		\item lag-specific directed links $\Xt \ix {t-\tau} \to \Xt \xj t$ pointing forward in time whenever $\Xt \ix {t-\tau}$ causes $\Xt \xj t$  with a time lag of $\tau > 0 $. 
	\end{itemize}
\end{definition} 
Note that lag-specific links may include self-transitions so long as $\tau > 0$.   

We make the standard assumptions of Causal Markov Condition, Faithfulness, and Sufficiency \cite{pearl:09:causalitybook}. 
\begin{assumption}[Causal Markov Condition, Faithfulness, and Sufficiency]  \label{asspt:causal} We assume the Causal Markov Condition and Faithfulness stating that $d$-separations in $\Gg$ correspond to conditional independencies in the data distribution $P$. We further assume that no unobserved confounding or selection variables exist. 
\end{assumption}

We also adopt a common assumption that the causal edge directions in the TCG persist over time.

\begin{assumption}[Repeating Edge Property~\cite{gerhardus:24:causalancestralgraphsts}]  
 We associate the time series $\XTDs $ with a  TCG  $\Gg$, with all edges remaining constant in direction across datasets $\D$ and through time $\T$, in the sense that if $\Xt \ix t \to \Xt \xj {t'}$ then $\Xt \ix {t+\tau} \to \Xt \xj {t'+\tau}$ is in $\Gg$ for any $\tau \geq 0$. 
\end{assumption}

Under this property of edge consistency over time, we can condense the information in the TCG into a single time window of the length of the maximum time lag $\taumax$.
\begin{definition}[Window Causal Graph~\cite{assad:22:ts_survey}] 
For the maximum time lag $\taumax$ in a TCG $\Gg$, the window causal graph (WCG) $\G =(V, E)$ over vertices $\Xt \ix t$ at each time   $t - \taumax, \ldots,  t$  contains  lag-specific directed links  $\Xt \ix {t-\tau} \to \Xt \xj t$ pointing forward in time whenever $\Xt \ix {t-\tau}$ causes $\Xt \xj t$ in $\Gg$ with time lag $\tau$, where $0 \leq \tau < \taumax$ if $\xj \neq \ix$ and  $0 < \tau < \taumax$ if $\xj = \ix$.  
\end{definition}

Under edge consistency, the causal mechanism changes preserve the causal parent sets. Their differences lie, therefore, in the structural mechanisms themselves as follows. 

\begin{assumption}[SCM with Contexts and Regimes]\label{asspt:regime-context-fcm}
	Given a time series 	$\XTDs$, 
    we assume a Structural Causal Model (SCM) $\FCM=(\Gg, \cps, \C, \R)$ with  TCG $\Gg$, changepoints $\cps$, and contexts and regimes $\C, \R$ for each $X$. Each variable $X$   is generated 
 through a set of structural equations of the form
	\begin{align} \label{eq:regime-context-fcm}
 \Xtc \ix t  d = \fcr \ix \ci \ri \big( \PA (\Xtc \ix t  d), \Ntc \ix t d \big) \end{align}
	where $\Xtc \ix t  d$ a sample  at time $t$ in set $d$ with $\textit{context}(d) = \ci$,  $\emph{regime}(t) = \ri$,  $\fcr \ix \ci \ri$ is its causal mechanism in $C_k$ and $R_r$, 
    $\PA (\Xtc \ix t  d)$ the set of its contemporaneous and lagged  parents in $\Gg$ 
	and $\Ntc \ix \ri \ci \indep \Xtc \ix  \ri \ci$ an independent noise variable.
\end{assumption}  
That is, all datasets and time points in a subsample  $(\ci, \ri)$ have identical generating process, where each variable has its mechanism $\fcr \ix \ci \ri$ and noise distribution $\Ntc \ix t \ci$. The above keeps variable indices implicit but can be similarly stated for all mechanisms $f_{(\ix)}$ of $X_{(\ix)}$  in the system. Note that our model can therefore show \emph{which} variables induce temporal or spatial changes at \emph{which} locations. 

Discovering this model from data will be our next focus.


\section{Causal Discovery with Contexts and Regimes}
Here, we describe our kernelized approach combined with minimum description lengths to discover the causal model underlying non-stationary time series.

\subsection{MDL for Causal Discovery} 
 
We base our approach on the Algorithmic Model of Causation (AMC)~\cite{janzing:08:amc}, which postulates that the factorization of the joint distribution with the lowest Kolmogorov complexity~\cite{kolmogorov} is the true causal model. In this paradigm, we reason about causal mechanisms as programs that compute effects from their parents. Kolmogorov complexity is not computable, but, can be approximated from above in a statistically sound manner via the Minimum Description Length (MDL)~\cite{rissanen:78:mdl, grunwald:07:book} principle. \citet{marx:21:formally} formally connect two-part MDL and the AMC.

MDL systematically balances model complexity and goodness of fit. 
The description length $L$ of the data $\mathbf X$ together with the model, here its causal graph $\Gg$, is given by $$L(\mathbf X ; \Gg) = \sum_{i \in \Gg}L(	\iX \ix\mid\PA(\iX \ix) ;  \M)\; $$ under a fixed model class $\M$.  
Each summand 
measures the length $L$, in bits, of both encoding the functional model class $\M$ itself as well as the data for $\iX \ix$ given $\PA(\iX \ix)$ under its optimal model. 
Our model class of choice are hereby Gaussian processes (GPs), also used by~\citet{mameche:23:linc}, as they are both non-parametric 
and have refined MDL scores~\citep{grunwald:07:book}, that is, there is a principled way of defining the model cost. 

A GP  models a distribution over functions
$f \sim \text{GP} (m(x), \kernel(x, x'))$
where $m(x) = \mathbb E [f(x)] $ is a mean function and
$
\kernel (x, x') = \mathbb E [(f(x)-m(x))(f(x')-m(x'))]
$ a  covariance kernel~\citep{gp-book:rasmussen}. 

The refined MDL description length~\cite{grunwald:07:book} of the data for a variable $\iX \ix $ given its causal parents $\PA(\iX \ix ) $ under its optimal GP model is given by 
\begin{align*}  
&L(	\iX \ix  \mid \PA(\iX \ix ) ;  \Mk) \nonumber
\\ &=
\min_{f \in \mathcal \Mk} \Big( -\log P(\iX \ix  \mid \PA (\iX \ix )) + \norm{f}^2_\kernel \Big) \nonumber \\  &+  \tfrac{1}{2} \log \det (\sigma^{-2} K_S + I)\;,
\end{align*}
where $K$ is the Gramian for $\kernel$ over $\PA(\iX \ix )$.

The score assesses the fit of the GP via the negative log-likelihood component, as well as its complexity via the squared norm $\norm{f}_{\kernel}^2 = \alpha^{\top} K \alpha$ of the space  $\M$ which can be seen as a measure of the smoothness of a given function, where $\alpha = (K + \sigma^2 I)^{-1} y$, 
and $\sigma^2$ is a scaling coefficient.  

\subsection{MDL for Non-stationary Time Series}
To encode our causal model, we consider the causal mechanisms per variable for each subsample under given contexts and regimes. We approximate each using a GP with additive Gaussian noise to obtain their description lengths.  

Given time series datasets, the MDL principle suggests the true causal model is the one minimizing this score,
\begin{align}\label{eq:objective} 
&\argmin_{\mathcal M} L(\XTDs; \mathcal M) \nonumber \\
=&\argmin_{\G, \cps, \big(\C, \R \big)}  \sum_{X \in \G}  \sum_{(\ri \in \R, \ci \in  \C)} L( \Xtc \ix \ri  \ci \mid \PA ( \Xtc \ix \ri  \ci ) ;  \Mk)
\;.
\end{align}
This raises the question of whether this score identifies the correct model components, including the WCG $\G$, the regime changepoint indices $\cps$, and partitions $\C$ and $\R$. 
 To this end, we impose a notion of persistence of regimes similar to the literature~\cite{saggioro:20:RPCMCI}.
 
\begin{assumption}[Sufficient Capacity and Persistence] \label{aspt:persistence}
	For each variable $X$ and subsample $ \W = (\ci, \ri) \in \C \times \R $, we have a large enough sample size $$|\W | = |\{ X_t^d \mid \emph{context}(d)=\ci, \emph{regime}(t)=k \}|$$ so that there exists a GP $ \fcr \ix \ci \ri  \in \M_{\kernel}$ such that Eq.~\eqref{eq:regime-context-fcm} holds.
	
	For each interval $w = [t_\text{min}, t_\text{max}] \subset \T$ subject to 
	\[ 
	\begin{cases} 
	\emph{regime}(t) = \ri \quad \forall t \in w \\
	\emph{regime}(t_\text{min}-1) \neq \ri \text{ and } \emph{regime}(t_\text{max}+1) \neq \ri\
	\end{cases}
	\]
	we have $|w| \geq d_\text{min}$, ensuring persistent regimes.
\end{assumption} 
 
It is important to note that time series are subject to auto-correlation and therefore not i.i.d., therefore we assume above that we access a large enough sample for each causal mechanism such that a GP can capture it. We also assume that each process persists for a minimal duration $d_{min}$ through time.
Furthermore,  we assume the following. 
\begin{assumption}[Shift Faithfulness]
\label{asspt:shiftfaith}
	  For distinct pairs $i \neq j, \ci \neq \ci'$ and $\ri \neq \ri'$ we have  $\ficr \ix \ci \ri \neq \ficr j {\ci'} {\ri'}$.
\end{assumption}  
Last, to allow us to distinguish causal directions we assume the principle of independent changes of causal mechanisms~\cite{towardscausalrep,perry:sms,coco}. 
We state it informally as follows and also express it through a latent-variable formulation of our causal model in the Appendix.
\begin{assumption}[Independent Changes~\cite{towardscausalrep}]
\label{asspt:indepchanges} The causal mechanisms  $f_{(\ix)}, f_{(j)}$ change independently of each other for each pair of variables $\iX \ix, \iX j$.  
\end{assumption}

Under these assumptions, we can show that our score is consistent, with the proof provided in the Appendix. 
  
\begin{theorem}[Consistency]
	\label{thm:2} 
Let Assumptions 1-6 hold. Then  Eq.~\eqref{eq:objective} is minimised for the true causal model $\FCM^\ast$ with WCG $\Gs, $ changepoints $\cps^{\ast}$, and partitions   ${\C}^{\ast}$ and $\R^{\ast}$ in the limit of $\D$ and $\T$.
\end{theorem} 

Given the large search space over models in Eq.~\eqref{eq:objective} and the inherent complexity in identifying the causal relationships and mechanisms while the regime changepoints and context and regimes partition are unknown, we develop a practical algorithm in the next section.


\begin{algorithm}[t]
	\caption{\ourmethod}
	\label{alg:ourmethod}
	\begin{algorithmic}[1]
        \REQUIRE Multivariate time series datasets $\XTDs$, minimum distance between changepoints $d_\emph{min}$, maximum time lag $\taumax$
        \ENSURE A window causal graph $\Gwindow$, a set of regime changepoints $\cps$, regime and context partitions $\R, \C$
        \STATE $\G = $ empty graph  
        \WHILE{not converged}
            \STATE $\cps = \textsc{ChangepointDetection} (\G, \XTDs, d_\emph{min}) $     \STATE $\R, \C = \textsc{Partitioning} (\G, \cps, \XTDs)$   
            \STATE $\Gwindow = $\textsc{EdgeGreedySearch}$(\cps, \R, \C, \XTDs, \taumax)$  
  
	\ENDWHILE
	\end{algorithmic}
\end{algorithm}  

\section{Algorithm}
Here, we introduce our algorithm \ourmethodeos\footnote{Code available at  \texttt{https://eda.group/spacetime}} for causal discovery over space (contexts) and time (regimes). Following our theory, we develop our algorithm around discovering the causal model that minimizes the summed description lengths in Eq.~\eqref{eq:objective}. Our algorithm's modular elements are therefore discovering the graph, changepoints, and partitions. 
We first describe how we discover a WCG  $\G$  for fixed $\cps, \C, $ and $\R$; then how we leverage changes of the functional models in  $\G$  to reveal contexts $\C$ and regimes $\R$; and last, how to identify the correct changepoints $\cps$  given $\G$. We iterate these steps until convergence.

\paragraph{Edge-Greedy Search}  MDL naturally allows ordering potential edges in the causal graph by their causal strength. Therefore, we adopt an edge-greedy search as proposed by~\citet{mian:21:globe}. Starting with an empty WCG $\G$ for a prespecified time lag $\taumax$, we first consider all pairwise edges among its nodes. 
  We define the edge strength of a directed edge  $E: \Xt \ix t \to \Xt j {t'}$ under the current model $\mathcal M$ as the relative gain in compression $$\delta E: = L(\XTDs; \mathcal M) - L(\XTDs ; \mathcal M')$$ where $\mathcal M'$ differs from $\mathcal M$ only by the additional edge $E$.

We proceed with a forward phase where we populate the graph  $\G$, adding edges in decreasing order of causal strength. Because this process is greedy, we may add edges that later become obsolete, so we follow it by a pruning phase. In this backward phase, we refine each variable's parent set and time lags based on our score and finally return the resulting WCG $\G$. 

\paragraph{Regime- and Context Partitioning} To discover regimes and contexts, we need to test for differences in causal mechanisms. Given two subsamples $\Wi = (d, w), \Wj =(d', w')$  for any time intervals $w, w'$ and $d, d' \in \D$,
we test for (in)equality in conditional distribution under our model, 
\[H_0: P(X
\mid \PA(X), \thetacr d w) \neq  P (X
\mid \PA(X),  \thetacr {d'} {w'}) \]
where $\PA(X)$ are the causal parents of $X$ in $\Gwindow$ and $\theta$ the GP parameters for the respective samples $\Wi, \Wj$. For this purpose, we apply kernelized hypothesis testing~\cite{cdnod,perry:sms}. Specifically, introducing a discrete variable $S$ that labels samples in $\Wi$ with $S=0$ resp. $\Wi'$ with $S=1$ we test the conditional independence $X \indep S \mid \PA(X)$ in the pooled data over $\Wi$ and $\Wi'$ using the kernel test HSIC~\cite{kci}.

We recall that we define the contexts in a spatial dimension and not constant over time. Therefore, we can perform this test pairwise over the datasets in $\D$, considering measurements aggregated over all of $\T$.
To arrive from pairwise tests to a partition $\C$, we group the datasets such that no group contains two datasets for which the test indicated a mechanism change.

Regime shifts, on the other hand, may occur over time irrespective of the context. Thus, we obtain the partition $\R$ by aggregating the pairwise tests performed over the time windows induced by $\cps$ with all datasets combined.

\paragraph{Changepoint Detection}
Unlike the dataset indices for the context partition, the regime changepoints must be identified beforehand. We propose an efficient method that takes advantage of the causal knowledge to identify changepoints without having to enumerate and test for mechanism change over all possible combinations over $\T$.
 
We recall that two regimes are distinguished by a change in the functional relationship between at least one variable and one of its parents. 
A functional model fitted on data from one regime should poorly predict data from another.
Hence, we identify the changepoints $\cps$ by looking for changes in the GPs prediction error over time, instead of directly in the raw data where it might not be obvious. 

We begin by fitting GPs to each variable and context within a time window where the regime is assumed constant, starting from the last known changepoint and of length equal to the minimum distance $d_\emph{min}$ between changepoints. 
We use the resultant GPs to predict the variable values across the entire dataset.
Then, we apply a statistical changepoint detection technique to the prediction errors across all variables and contexts simultaneously, identifying the next changepoint. Many different techniques can be used here; a good choice is PELT~\cite{pelt}, an efficient algorithm that does not require the number of changepoints.

\paragraph{Iterative Model Learning}
We apply an iterative strategy to combine the above steps, as summarized in Alg.~\ref{alg:ourmethod}.
Initially, we segment the temporal space $\T$ into changepoints under an empty graph (Line 1), given that this captures marginal distribution shifts.  
An iteration consists of detecting changepoints $\cps$ (Line 3) using changes in residual distribution of the functions $\fcr \ix \ci \ri$ under the current graph; learning the regime and context partitions (Line 4) using the current graph and updated changepoints; and finally, 
discovering $\G$ (Line 5) with a greedy strategy, where we need the current changepoints and partitions to obtain the scores in Eq.~\eqref{eq:objective}. We repeat these steps until convergence. Pseudo-codes of the components are deferred to the Appendix.


\section{Experiments} 

\def\ns{11}
\def\nsgpcps{11}
\def\nsgpkndagcps{11}
\def\nsrpcmci{11}
\def\nsrpcmcicps{11}
\def\nsvlng{6}
\def\nsvlngkncps{6}
\def\nonlinear{expres/box/c_2_r_3_cps_2_n_5_t_200_d_1_i_0.5_ivs_50_f_exp_ns_gauss_hi_False} 
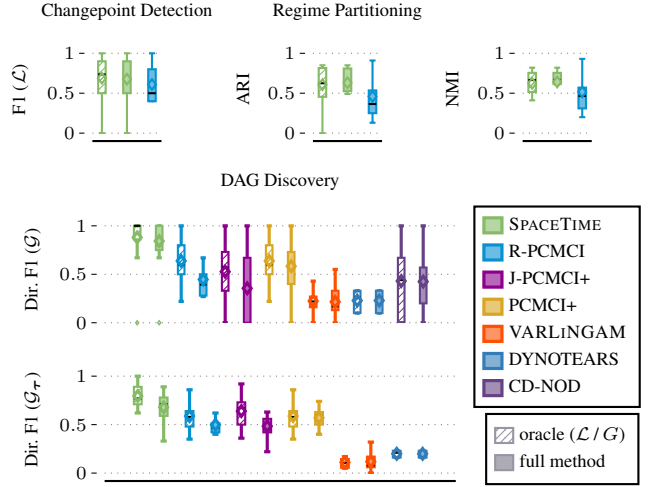
\def\id{\emph{ \renewcommand{\cps}{\mathcal L} Causal Discovery, Changepoint Detection and Regime Partitioning}. We evaluate the methods on multiple time series with causal mechanism shifts across time and datasets, with non-linear functional form, Gaussian noise, where $|\G|=5, |\T| = 200, |\C| =2, |\R|  = 3, |\mathcal L| = 2,$ and a fraction $s=\frac{1}{2}$ of intervened edges. }
\def\oracles{gp-kncps/\ns,  R-pcmci-kncps/\ns, pcmciplus/\ns,  
varLINGAM-kncps/\nsvlngkncps, dynotears-kncps/\ns, cdnod-kncps/\ns} 
\def\timedoracles{gp-kncps/\ns, J-pcmci-kncps/\ns, R-pcmci-kncps/\ns, R-pcmci-kncps/\ns, 
varLINGAM-kncps/\nsvlngkncps, dynotears-kncps/\ns} 
\def\cps{gp-kndag/\ns, gp/\ns, R-pcmci/\nsrpcmci}
\def\cpsadjusted{gp-kndag/\nsgpkndagcps, gp/\nsgpcps, R-pcmci/\nsrpcmcicps}
\def\methods{gp-kncps/\ns,gp/\ns, 
J-pcmci-kncps/\ns, J-pcmci/\ns, R-pcmci-kncps/\ns, R-pcmci/\nsrpcmci,  R-pcmci-kncps/\ns, 
pcmciplus/\ns, varLINGAM-kncps/\nsvlngkncps, varLINGAM/\ns,  dynotears-kncps/\ns, dynotears/\ns, cdnod-kncps/\ns, cdnod/\ns}
\def\timedmethods{gp-kncps/\ns,gp/\ns, 
J-pcmci-kncps/\ns, J-pcmci/\ns, R-pcmci-kncps/\ns, R-pcmci/\nsrpcmci,  R-pcmci-kncps/\ns, 
pcmciplus/\ns, varLINGAM-kncps/\nsvlngkncps, varLINGAM/\ns,  dynotears-kncps/\ns, dynotears/\ns} 
\begin{figure} [t] 
 \begin{minipage}{0.8\textwidth}
 \begin{tikzpicture} 
		\begin{axis}[ 
			pretty boxplot regime, title={\scriptsize{Changepoint Detection }},  
			 pretty enlargexlimits,  
			scatter src			= explicit symbolic,   
			pretty xlabelrot, ymin = 0, ymax= 1.1, width=2.5cm,height = 2.75cm,
			ylabel={F1 ($\mathcal L$)}, xlabel={ }, xticklabels=none,  pretty labelshift,
			] 
			\def\metric{cps-f1}; 
			\def\file{\nonlinear};  
			\foreach \name/\succ in \cps{   
			\addplot+[area legend, boxplot = {%
				average = auto,%
				every average/.style={/tikz/mark=diamond*,/tikz/mark size =2pt},%
			}, 
			]  table[y=\name_\metric] {\file/successes_\succ/\name.csv};   
			}   
		\end{axis}
	\end{tikzpicture}  
   \begin{tikzpicture} 
		\begin{axis}[ 
			pretty boxplot regime, title={\scriptsize{Regime Partitioning}},  
			pretty enlargexlimits,  
			scatter src			= explicit symbolic,   
			pretty xlabelrot, ymin = 0, ymax= 1.1,width=2.5cm,height = 2.75cm,
			ylabel={ARI}, xlabel={ }, xticklabels=none, 		legend entries={ }, legend columns=1, legend style=  {at = {(1.1, 2)}, anchor	= north west, font=\scriptsize},  
	 	pretty labelshift,
			] 
			\def\metric{cps-ari}; 
			\def\file{\nonlinear};  
			\foreach \name/\succ in \cps{   
				\addplot+[ area legend, boxplot = {%
					average = auto,
					every average/.style={/tikz/mark=diamond*,/tikz/mark size =2pt},%
					 }]  table[y=\name_\metric] {\file/successes_\succ/\name.csv};   
			}     
		\end{axis}
	\end{tikzpicture} 
 		\begin{tikzpicture} 
		\begin{axis}[ 
			pretty boxplot regime, title={  },  
			pretty enlargexlimits,  
			scatter src			= explicit symbolic,   width=2.5cm, height = 2.75cm,
			pretty xlabelrot, ymin = 0, ymax= 1.1,
			ylabel={NMI}, xlabel={ }, xticklabels=none,
   legend entries={ 
   }, legend style = {at={(3, 1.6)}},legend image post style={xscale=.4, yscale=1}, 
	 	pretty labelshift,
			] 
			\def\metric{cps-nmi}; 
			\def\file{\nonlinear};  
			\foreach \name/\succ in \cps{   
				\addplot+[ area legend, boxplot = {%
					average = auto,
					every average/.style={/tikz/mark=diamond*,/tikz/mark size =2pt},%
					 }]  table[y=\name_\metric] {\file/successes_\succ/\name.csv};   
			}     
		\end{axis}  
	\end{tikzpicture}   
 \end{minipage}
    
\def\ns{20}
\def\nsgpcps{12}
\def\nsgpkndagcps{13}
\def\nsrpcmci{5}
\def\nsrpcmcicps{0}
\def\nsvlng{20}
\def\nsvlngkncps{11}
\def\nonlinear{expres/box/c_2_r_3_cps_2_n_5_t_200_d_1_i_0.5_ivs_50_f_exp_ns_gauss_hi_False_2} 
\def\oracles{gp-kncps/\ns,  R-pcmci-kncps/\ns, pcmciplus/\ns,  
varLINGAM-kncps/\nsvlngkncps, dynotears-kncps/\ns, cdnod-kncps/\ns} 
\def\timedoracles{gp-kncps/\ns, R-pcmci-kncps/\ns, J-pcmci-kncps/\ns, R-pcmci-kncps/\ns, 
varLINGAM-kncps/\nsvlngkncps, dynotears-kncps/\ns} 
\def\cps{gp-kndag/\ns, gp/\ns, R-pcmci/\nsrpcmci}
\def\cpsadjusted{gp-kndag/\nsgpkndagcps, gp/\nsgpcps, R-pcmci/\nsrpcmcicps}
\def\methods{gp-kncps/\ns,gp/\ns,  R-pcmci-kncps/\ns, R-pcmci/\nsrpcmci,  
J-pcmci-kncps/\ns, J-pcmci/\ns,R-pcmci-kncps/\ns, 
pcmciplus/\ns, varLINGAM-kncps/\nsvlngkncps, varLINGAM/\ns,  dynotears-kncps/\ns, dynotears/\ns, cdnod-kncps/\ns, cdnod/\ns}
\def\timedmethods{gp-kncps/\ns,gp/\ns,  R-pcmci-kncps/\ns, R-pcmci/\nsrpcmci, 
J-pcmci-kncps/\ns, J-pcmci/\ns, R-pcmci-kncps/\ns, 
pcmciplus/\ns, varLINGAM-kncps/\nsvlngkncps, varLINGAM/\ns,  dynotears-kncps/\ns, dynotears/\ns} 
	\centering 
 \begin{minipage}{0.45\textwidth}
	\begin{tikzpicture}  
		\begin{axis}[
			pretty boxplot regime, title={\scriptsize{DAG Discovery}},  
			pretty enlargexlimits,  
			scatter src			= explicit symbolic,  
			cycle list name =prcl-box-time,
			pretty xlabelrot, 
   axis x line = none, x axis line style={draw opacity=0},
			ymin = 0,  ymax= 1.1, 
   width= 6.25cm, height=3cm,
			ylabel={Dir. F1 ($\Gg$)}, xlabel={ }, xticklabels=none, legend entries={, \ourmethod,,R\nobreakdash-PCMCI,,J\nobreakdash-PCMCI+,,
   PCMCI+,, \textsc{VARLiNGAM},, DYNOTEARS, ,CD\nobreakdash-NOD},legend columns=1, legend style=  {at = {(1.05, 1.1)}, anchor	= north west, font=\scriptsize},  legend image post style={xscale=.4, yscale=1}, 
			pretty labelshift,
			] 
			\def\metric{f1};
			\def\var{C}; 
			\def\file{\nonlinear};  
			\foreach \name/\succ in \methods{   
				\addplot+[ area legend, boxplot = {%
				average = auto,
				every average/.style={/tikz/mark=diamond*,/tikz/mark size =2pt},%
			}] table[y=\name_\metric] {\file/successes_\succ/\name.csv};   
			}     
		\end{axis}
       \begin{scope}[yshift=-0.25\linewidth] 
 \begin{axis}[pretty scatter, height = 4.25cm, width = 5cm,   	ymin = -8, ymax = 30, xmin = 0, xmax = 150,  axis line style={draw=none}, tick style={draw=none},  legend style = {at={(2.08, .3)}},  legend image post style={xscale=.4, yscale=1}, xlabel = { },ylabel = { }, yticklabels = {},xticklabels = {}, pretty labelshift ] 
 \addlegendimage{area legend,  lilacgray,  fill=lilacgray!20 , pattern=north east lines, pattern color=lilacgray  }
 \addlegendentry{\textcolor{black!60}{}oracle ($\mathcal L\,$/$\,G$)}\addlegendimage{area legend,  lilacgray,  fill=lilacgray!80}\addlegendentry{\textcolor{black!60}{}full method} \end{axis} 
		\begin{axis}[ 
			pretty boxplot regime, 
			pretty enlargexlimits,  
			scatter src			= explicit symbolic,  
			cycle list name =prcl-box-time,
			pretty xlabelrot, 
			ymin = 0, ymax= 1.1, width= 6.25cm, height=3cm,
			ylabel={Dir. F1 ($\G$)}, xlabel={ }, xticklabels=none,
			pretty labelshift,
			] 
			\def\metric{f1-timed};
			\def\var{C}; 
			\def\file{\nonlinear};  
			\foreach \name/\succ in \timedmethods{   
				\addplot+[ area legend, boxplot = {%
					average = auto,
					every average/.style={/tikz/mark=diamond*,/tikz/mark size =2pt},%
				}] table[y=\name_\metric] {\file/successes_\succ/\name.csv};   
			}     
    \addplot+[area legend, boxplot = {%
				average = auto,%
				every average/.style={/tikz/mark=diamond*,/tikz/mark size =2pt},%
			}, 
			] coordinates {(0,-1)};
      \addplot+[area legend, boxplot = {%
				average = auto,%
				every average/.style={/tikz/mark=diamond*,/tikz/mark size =2pt},%
			}, 
			] coordinates {(0,-1)}; 
		\end{axis} 
	\end{scope}
	\end{tikzpicture}
 \end{minipage}

 \caption{ 	\label{fig:main_synthetic_experiment} \id }
 \end{figure}
 \renewcommand{\cps}{\mathcal L}

In this section, we evaluate our approach on synthetic data and real-world datasets in hydrology and meteorology. 
 
\subsection{Experimental Setup}
To simulate time series datasets, we generate random WCGs $\G$ with maximum lag $\taumax=2$, with corresponding summary DAG $\Gg$. 
We sample regime changepoints $\cps$ uniformly at random using a pre-set minimal duration, set to $d_\emph{min}=30$ unless otherwise specified. In each context and regime, we re-sample the edge weights of a fraction of all edges, and finally sample data similarly to~\citet{gunther:23:jpcmcip} using \textsc{Tigramite}~\citet{runge:20:pcmcip}, with linear or non-linear functional form and Gaussian or uniform noise. 

Our first real-world case study investigates the effect of meteorological drivers on river discharge over different gauged catchments across Europe, given data derived from the Global Runoff Data Centre (GRDC) datasets~\cite{cornes:18:riverdata}. Secondly, we study  biosphere–atmosphere fluxes in the FLUXNET dataset~\cite{baldocchi:20:fluxdata}.
Both cases include multiple datasets $\D$ from diverse geographical locations where the respective time series span multiple years, with daily time resolution. After pre-processing~\cite{guenther:23:river,krich:21:flux}, we obtain $|\D| = 307$ locations for the river runoff, resp. $|\D| = 63$ locations for the fluxes data, and set $\T$ to a period of one year (2006 resp. 2010) during DAG discovery given that the available years are non-overlapping.  
We  provide the datasets $\D$ from all locations to our method, using fixed hyperparameters $d_\emph{min}=30$ and $\taumax=2$ for consistency.

\begin{figure}[t]
    \input{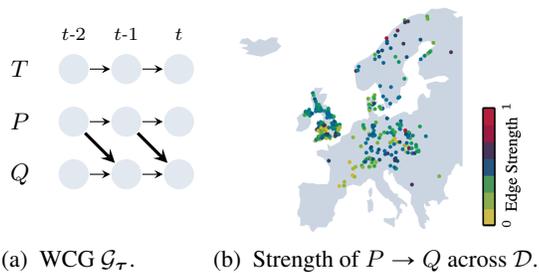}
    \caption{ \emph{Drivers of River Discharge}. Given temperature (\emph{T}), precipitation (\emph{P}) and river discharge (\emph{Q}) in 307 catchments across Europe, we discover a common WCG $\G$ for $\taumax=2$ (a) with a lagged edge  $P\to \emph{Q}$, and illustrate its variability in causal strength between catchments (b).  
    }
    \label{fig:river}
\end{figure}

Throughout our experiments, we are interested in how the modular components of \ourmethod work together, which leads to the following questions: whether it discovers causal directions correctly over time and contexts; whether it accurately detects changepoints over time; and whether it finds meaningful partitions of similar causal mechanisms.

\subsection{Synthetic Data}
 
We show our results on synthetic data in Fig.~\ref{fig:main_synthetic_experiment}.  We report F1 scores over $\cps$  (top) for changepoint detection, F1 scores over directed edges in $\G$ (middle) and the DAG $\Gg$ (bottom) for causal discovery, and the Adjusted Rand Index (ARI) and Normalized Mutual Information (NMI) for regime partitioning.  
We also report changepoint detection with known $\G$ resp. causal discovery with known $\cps$ (hatched). 
How we apply the methods depends on whether they can handle multiple datasets (ours, JPCMCI) resp. multiple regimes (ours, \rpcmci). We apply methods that assume a single dataset to each $d$ in turn. Across time, we apply methods that assume a single regime either to each subsample $s$ with known  $\cps$ (hatched), or all regimes together (solid). We aggregate all resulting graphs using a majority vote for each edge in the graph.
We find that \ourmethod consistently performs well compared to its competitors, exemplified in Fig.~\ref{fig:main_synthetic_experiment} for the non-linear Gaussian case, and with varying functional form and noise distributions which we postpone to the Appendix.

\paragraph{Causal Discovery} Regarding causal discovery, the methods \dynotears and \varlingam make linearity assumptions and the latter in addition assumes that the noises in the model are non-Gaussian. Therefore, it is not surprising that they show sub-par performance here. 
Even though they and  \pcmcip assume stationarity, i.e. the absence of regime shifts, the effect of known (hatched) compared to unknown changepoints (solid) is not overly pronounced. This could be due to  a slight advantage due to increased sample size when combining all regimes and the fact that changes in edge strength are not as drastic as changes in the graph.  
While \cdnod can address temporal shifts, it only discovers a summary DAG $\Gg$.  
As all the aforementioned methods assume a single time series, note that we apply them to each dataset and combine the resulting graphs using majority voting. 
Only \jcmcip can consider multiple contexts, but does not achieve a significant performance gap here. 
This may be due to mechanism shifts being too subtle, given that \jcmcip relies on changing graphs across contexts,  
or because it may need a larger number of different contexts to effectively outperform single-context methods, in which case conditional independence tests  however become increasingly expensive due to a high dimensional one-hot encoding. 

\paragraph{Changepoint and Regime Identification} In the changepoint detection and partitioning experiments (Fig.~\ref{fig:main_synthetic_experiment} top), we can see that \ourmethod is robust to a small number of mistakes in the causal graph as the full results (solid) are close to the oracle version (hatched). This also matches our observation that the interleaving steps in Alg.~\ref{alg:ourmethod} converge quickly, typically within three iterations.  
Reasons that we outperform \rpcmci (Fig.~\ref{fig:main_synthetic_experiment}) may include the fact that we impose regime persistence explicitly, whereas \rpcmci can only impose a given number of regime-switches here provided as background knowledge.

\subsection{Case Study: Identifying Drivers of River Discharge}
As a real-world application, we explore the effects of meteorological variables on river discharge using data from so-called catchments, that is, an area where water drains into an outlet.  
We focus on temperature (\emph{T}), precipitation (\emph{P}) as well as river runoff (\emph{Q}) in our analysis. 

\tikzset{ 
    dag_node/.style={draw=none, thick, rounded corners, circle,
        minimum size=0.4cm, fill=slate200
        }
}

\def\timeseriesA{5_6943100_2010.csv} 
\def\cpsAone{71}
\def\cpsAtwo{115}
\def\cpsAthree{145}
\def\cpsAfour{198}
\def\cpsAfive{243} 
\def\cpsAsix{301}

\def\timeseriesB{5_6458406_2010.csv} 
\def\cpsBone{79}
\def\cpsBtwo{126}
\def\cpsBthree{176}
\def\cpsBfour{206}
\def\cpsBfive{284} 
\def\cpsBsix{284}

\def\timeseriesC{6_6604220_2010.csv}     
\def\cpsCone{46}
\def\cpsCtwo{77}
\def\cpsCthree{111}
\def\cpsCfour{190}
\def\cpsCfive{220}
\def\cpsCsix{255}
\def\cpsCseven{301}
\def\cpsCeight{332}

\def\timeseriesCc{6_6457320_2010.csv} 
\def\timeseriesBb{6_6458924_2010.csv} 
\def\timeseriesAa{5_6243030_2010.csv}
\def\cpsAaone{115}
\def\cpsAatwo{145}
\def\cpsAathree{199}
\def\cpsAafour{254}
\def\cpsAafive{288} 
\def\cpsAasix{288}

\def\mix{6_6939500_2010.csv} 
\def\timeseriesused{6_6604220_2010.csv} 
\def\nicemix{6_6233250_2010.csv} 
\def\niceperiodic{5_6943100_2010.csv} 
\def\nicespikes{5_6458406_2010.csv} 
\def\nicemountain{5_6243030_2010.csv} 
\def\smallspikes{3_6934460_2010.csv} 
\def\spiky{3_6604180_2010.csv} 
\def\mountain{5_6935145_2010.csv} 
\def\yshiftts{20}
\def\poscontextts{30}
\def\poslabelts{\poscontextts+30}
\def\xshiftts{\poslabelts+30}
\def\colorregimezero{darkmagenta}
\def\colorregimeone{mambacolor4}
\def\colorregimetwo{dollarbill}
\def\colorregimethree{goldenrod}
\def\colorregimefour{orange}
\def\colorregimefive{niceblue}
\def\colorcontextone{mambacolor4}
\def\colorcontexttwo{nipsvio}

\begin{figure}[t] 
        \centering
        \begin{tikzpicture}
        \begin{axis} [
            scale only axis,
            width=\linewidth,
            height=1.25cm,
            axis x line=none,  
            axis y line=none,  
            xtick=\empty,
            ytick=\empty,
            xmin=0,
            legend entries={}, 
            legend image post style={xscale=.4, yscale=1}, 
            legend style={font=\tiny},
        ] 

        \node[draw=none] at (axis cs:\poscontextts,-\yshiftts) {$\big(d_1\big)$}; 
        \node[draw=none] at (axis cs:\poslabelts,0) {\emph{Q}}; 
        
        \addplot[\colorregimezero, mark=none, unbounded coords=jump, x filter/.expression={x>\cpsAone ? nan : x+\xshiftts}
        ] table [col sep=comma, x=t, y=Qobs] {expres/river_texdata/\timeseriesA};%
 
        \addplot[\colorregimezero, mark=none, unbounded coords=jump, x filter/.expression={x<\cpsAsix ? nan : x+\xshiftts}] table [col sep=comma, x=t, y=Qobs]  {expres/river_texdata/\timeseriesA}; 

        \addplot[\colorregimeone, mark=none, unbounded coords=jump, x filter/.expression={x<\cpsAone || x>\cpsAtwo ? nan : x+\xshiftts}] table [col sep=comma, x=t, y=Qobs]  {expres/river_texdata/\timeseriesA};

        \addplot[\colorregimetwo, mark=none, unbounded coords=jump, x filter/.expression={x<\cpsAtwo || x>\cpsAthree ? nan : x+\xshiftts}] table [col sep=comma, x=t, y=Qobs]  {expres/river_texdata/\timeseriesA};        
        \addplot[\colorregimetwo, mark=none, unbounded coords=jump, x filter/.expression={x<\cpsAfour || x>\cpsAfive ? nan : x+\xshiftts}] table [col sep=comma, x=t, y=Qobs]  {expres/river_texdata/\timeseriesA};  
        
        \addplot[\colorregimethree, mark=none, unbounded coords=jump, x filter/.expression={x<\cpsAthree || x>\cpsAfour ? nan : x+\xshiftts}] table [col sep=comma, x=t, y=Qobs]  {expres/river_texdata/\timeseriesA}; 
        \addplot[\colorregimefour, mark=none, unbounded coords=jump, x filter/.expression={x<\cpsAfive || x>\cpsAsix ? nan : x+\xshiftts}] table [col sep=comma, x=t, y=Qobs]  {expres/river_texdata/\timeseriesA};

        \def\ytimeline{-5*\yshiftts}
        \addplot[draw=none] coordinates{(\xshiftts,\ytimeline) (\xshiftts+100,\ytimeline-3)};
        \draw[->] (axis cs:\xshiftts,\ytimeline) -- (axis cs:\xshiftts+365,\ytimeline);
        \addplot[mark=|] coordinates{(\xshiftts+\cpsAone,\ytimeline) (\xshiftts+\cpsAtwo,\ytimeline)};
        \addplot[mark=|] coordinates{(\xshiftts+\cpsAthree,\ytimeline) (\xshiftts+\cpsAfour,\ytimeline)};
            \addplot[mark=|] coordinates{(\xshiftts+\cpsAfive,\ytimeline) (\xshiftts+\cpsAsix,\ytimeline)}; 
        \draw[->, thin] (axis cs:\xshiftts,\ytimeline) -- (axis cs:\xshiftts+365,\ytimeline);
        \node[draw=none] at (axis cs:\xshiftts+370,\ytimeline) {$t$};
        \end{axis} 
       \end{tikzpicture} 
        \begin{tikzpicture}
        \begin{axis} [
            scale only axis,
            width=\linewidth,
            height=1.25cm,
            axis x line=none,  
            axis y line=none,  
            xtick=\empty,
            ytick=\empty,
            xmin=0,
            legend entries={}, 
            legend image post style={xscale=.4, yscale=1}, 
            legend style={font=\tiny},
        ] 
        \node[draw=none] at (axis cs:\poscontextts,-\yshiftts) {$\big(d_2\big)$}; 
        \node[draw=none] at (axis cs:\poslabelts,0) {\emph{Q}}; 
        \addplot[\colorregimezero, mark=none, unbounded coords=jump, x filter/.expression={x>\cpsBone ? nan : x+\xshiftts}
        ] table [col sep=comma, x=t, y=Qobs] {expres/river_texdata/\timeseriesB};%

        \addplot[\colorregimezero, mark=none, unbounded coords=jump, x filter/.expression={x<\cpsBfive ? nan : x+\xshiftts}] table [col sep=comma, x=t, y=Qobs]  {expres/river_texdata/\timeseriesB}; 
         
        \addplot[\colorregimeone, mark=none, unbounded coords=jump, x filter/.expression={x<\cpsBone || x>\cpsBtwo ? nan : x+\xshiftts}] table [col sep=comma, x=t, y=Qobs]  {expres/river_texdata/\timeseriesB};

        \addplot[\colorregimetwo, mark=none, unbounded coords=jump, x filter/.expression={x<\cpsBtwo || x>\cpsBthree ? nan : x+\xshiftts}] table [col sep=comma, x=t, y=Qobs]  {expres/river_texdata/\timeseriesB};

        \addplot[\colorregimethree, mark=none, unbounded coords=jump, x filter/.expression={x<\cpsBthree || x>\cpsBfour ? nan : x+\xshiftts}] table [col sep=comma, x=t, y=Qobs]  {expres/river_texdata/\timeseriesB}; 
        
        \addplot[\colorregimefour, mark=none, unbounded coords=jump, x filter/.expression={x<\cpsBfour || x>\cpsBfive ? nan : x+\xshiftts}] table [col sep=comma, x=t, y=Qobs]  {expres/river_texdata/\timeseriesB};

        \def\ytimeline{-15*\yshiftts}
        \addplot[draw=none] coordinates{(\xshiftts,\ytimeline) (\xshiftts+100,\ytimeline-3)};
        \draw[->] (axis cs:\xshiftts,\ytimeline) -- (axis cs:\xshiftts+365,\ytimeline);
        \addplot[mark=|] coordinates{(\xshiftts+\cpsBone,\ytimeline) (\xshiftts+\cpsBtwo,\ytimeline)};
        \addplot[mark=|] coordinates{(\xshiftts+\cpsBthree,\ytimeline) (\xshiftts+\cpsBfour,\ytimeline)};
            \addplot[mark=|] coordinates{(\xshiftts+\cpsBfive,\ytimeline) (\xshiftts+\cpsBsix,\ytimeline)}; 
        \draw[->, thin] (axis cs:\xshiftts,\ytimeline) -- (axis cs:\xshiftts+365,\ytimeline);
        \node[draw=none] at (axis cs:\xshiftts+370,\ytimeline) {$t$};
        \end{axis} 
       \end{tikzpicture}
        \begin{tikzpicture}
        \begin{axis} [
            scale only axis,
            width=\linewidth,
            height=1.25cm,
            axis x line=none,  
            axis y line=none,  
            xtick=\empty,
            ytick=\empty,
            xmin=0,
            legend entries={}, 
            legend image post style={xscale=.4, yscale=1}, 
            legend style={font=\tiny},
        ] 
        \node[draw=none] at (axis cs:\poscontextts,-0.2*\yshiftts) {$\big(d_3\big)$}; 
        \node[draw=none] at (axis cs:\poslabelts,0) {\emph{Q}}; 
        \addplot[\colorregimezero, mark=none, unbounded coords=jump, x filter/.expression={x>\cpsCone ? nan : x+\xshiftts}
        ] table [col sep=comma, x=t, y=Qobs] {expres/river_texdata/\timeseriesC};%

        \addplot[\colorregimeone, mark=none, unbounded coords=jump, x filter/.expression={x<\cpsCone || x>\cpsCtwo ? nan : x+\xshiftts}] table [col sep=comma, x=t, y=Qobs]  {expres/river_texdata/\timeseriesC};
        
        \addplot[\colorregimeone, mark=none, unbounded coords=jump, x filter/.expression={x<\cpsCthree || x>\cpsCfour ? nan : x+\xshiftts}] table [col sep=comma, x=t, y=Qobs]  {expres/river_texdata/\timeseriesC}; 
        
        \addplot[\colorregimeone, mark=none, unbounded coords=jump, x filter/.expression={x<\cpsCfive|| x>\cpsCsix ? nan : x+\xshiftts}] table [col sep=comma, x=t, y=Qobs]  {expres/river_texdata/\timeseriesC}; 
        
        \addplot[\colorregimetwo, mark=none, unbounded coords=jump, x filter/.expression={x<\cpsCtwo || x>\cpsCthree ? nan : x+\xshiftts}] table [col sep=comma, x=t, y=Qobs]  {expres/river_texdata/\timeseriesC};
        
        \addplot[\colorregimetwo, mark=none, unbounded coords=jump, x filter/.expression={x<\cpsCsix || x>\cpsCseven? nan : x+\xshiftts}] table [col sep=comma, x=t, y=Qobs]  {expres/river_texdata/\timeseriesC}; 
        
        \addplot[\colorregimethree, mark=none, unbounded coords=jump, x filter/.expression={x<\cpsCfour || x>\cpsCfive ? nan : x+\xshiftts}] table [col sep=comma, x=t, y=Qobs]  {expres/river_texdata/\timeseriesC}; 
        \addplot[\colorregimefour, mark=none, unbounded coords=jump, x filter/.expression={x<\cpsCseven || x>\cpsCeight  ? nan : x+\xshiftts}] table [col sep=comma, x=t, y=Qobs]  {expres/river_texdata/\timeseriesC}; 
        \addplot[\colorregimefive, mark=none, unbounded coords=jump, x filter/.expression={x<\cpsCeight ? nan : x+\xshiftts}] table [col sep=comma, x=t, y=Qobs]  {expres/river_texdata/\timeseriesC};

        \def\ytimeline{-0.75*\yshiftts}
        \addplot[draw=none] coordinates{(\xshiftts,\ytimeline) (\xshiftts+100,\ytimeline-3)};
        \draw[->] (axis cs:\xshiftts,\ytimeline) -- (axis cs:\xshiftts+365,\ytimeline);
        \addplot[mark=|] coordinates{(\xshiftts+\cpsCone,\ytimeline) (\xshiftts+\cpsCtwo,\ytimeline)};
        \addplot[mark=|] coordinates{(\xshiftts+\cpsCthree,\ytimeline) (\xshiftts+\cpsCfour,\ytimeline)};
            \addplot[mark=|] coordinates{(\xshiftts+\cpsCfive,\ytimeline) (\xshiftts+\cpsCsix,\ytimeline)};
            \addplot[mark=|] coordinates{(\xshiftts+\cpsCseven,\ytimeline) (\xshiftts+\cpsCeight,\ytimeline)};
        \draw[->, thin] (axis cs:\xshiftts,\ytimeline) -- (axis cs:\xshiftts+365,\ytimeline);
        \node[draw=none] at (axis cs:\xshiftts+370,\ytimeline) {$t$};
        \end{axis} 
       \end{tikzpicture}
\caption{\emph{Changepoints of the Interaction between Precipitation and Runoff.} We show the regime changepoints $\cps$ and partition $\R$ that we discover for  $P \to \emph{Q}$ with \ourmethod for selected catchments $d_1$ (CH), $d_2$ (PL), $d_3$ (GB). 
The colors denote different regimes.}
    \label{fig:river_regimes}
\end{figure}
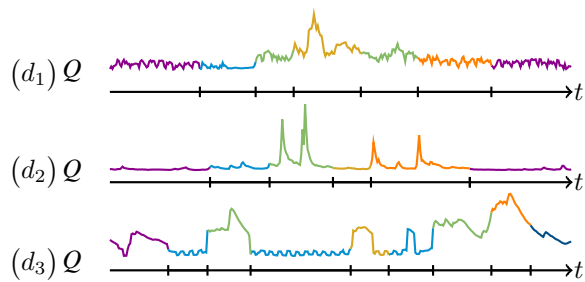
 
\paragraph{Causal Discovery over Different Catchments} While a previous study of this data using \pcmcip was limited in considering each location and month separately \cite{guenther:23:river}, we are interested in a unified analysis of all catchments.  
As we show in Fig.~\ref{fig:river_a}, we discover a direct influence of precipitation on discharge through the lagged edge $E: P_{(t-1)} \to \emph{Q}_{(t)}$ which we consider reasonable. 
The joint graph aside, we also visualize the causal strength $\delta E$ of this edge for each of the 307 locations in the year 2010 in Fig.~\ref{fig:river_b}. We observe regional similarities but also geographical variation in causal strength. The latter is likely due to differences in topology and climate in the different parts of Europe, plus distinct characteristics of each catchment such as area, altitude, and vegetation~\cite{guenther:23:river}.
 
\paragraph{Changepoints and Regimes over Time}  Fig.~\ref{fig:river_regimes} shows  for selected examples  the changepoints and regimes with similar causal relationships that \ourmethod discovers. 
 
\pgfplotsset{colormap={ourmap}{
rgb255=(251,191,36) rgb255=(22,161,74) rgb255=(3,105,161) rgb255=(131,24,76)}%
}
\begin{figure}[t] 
    \begin{subfigure}[t]{0.2\textwidth}
     \begin{tikzpicture}[->,>=stealth, node distance=1cm, every node/.style={draw, thick, font=\small}] 
		\foreach[count=\i,evaluate=\i as \angle using (\i)*360/6] \text in {%
			VPD,LE,H,NEE,R,T
		}
		\node[] (n\i) [dag_node, fill=slate200] at (\angle:1.1) {};
  \node[invis_node, below=of n5] {}; 
		\draw [dag_edge] (n5) edge (n3); 
		\draw [dag_edge] (n2) edge (n5); 
		\draw [dag_edge] (n5) edge (n4); 
		\draw [dag_edge] (n5) edge (n1); 
		\draw [dag_edge] (n3) edge (n2); 
		\draw [dag_edge] (n1) edge (n2);   
		\draw [dag_edge] (n1) edge (n6);  

every node/.style={thick, font=\small}] 
		\foreach[count=\i,evaluate=\i as \angle using (\i)*360/6] \text in {%
			VPD,LE,H,NEE,$R_g$,T
		}
		\node[] (n\i) [draw=none] at (\angle:1.55) {\emph{\text } };  

        \end{tikzpicture}
        \caption{ \label{fig:flux_a}
        Causal DAG $\Gg$.}
    \end{subfigure} 
 \begin{subfigure}[t]{.24\textwidth} 
 \centering
\begin{tikzpicture}[square/.style={regular polygon,regular polygon sides=4}]
    \begin{axis}[pretty heated, height=4.5cm, width=4.5cm,
        colorbar,
        mesh/interior colormap name=hot, 
        colormap={ourmap}{rgb255=(251,191,36) rgb255=(22,161,74) rgb255=(3,105,161) rgb255=(131,24,76)},
            scatter/use mapped color={draw=mapped color!95, fill=mapped color!65},
        colorbar sampled,
        colorbar style={samples=6,
            ytick={0,  1}, 
            yticklabels={0,1},
            width=0.2cm, height=1.4cm, yshift=-1.5cm,
            xshift=-0.1cm, 
            ticklabel style={
                /pgf/number format/fixed,    
                /pgf/number format/precision=2, 
                font=\scriptsize,        
                ylabel = {Edge strength 
                    }, 
            }, 
            yticklabels={0,1},
            xticklabels={},
            yticklabel style={
                font=\tiny,  text width=.5em,
                align=right,
                /pgf/number format/.cd,
                fixed,
                fixed zerofill
            }
        }, 
        mark size=1pt,
        xlabel={$t$-SNE $x_1$}, 
        ylabel={$t$-SNE $x_2$}, xmax=140, xmin =-140, ymax=120, ymin=-120,  
        pretty spacy, 
        ] 
            \addplot+[
            scatter, line width =0.2pt,
            point meta=explicit,  
        ] table[meta=col] {expres/realworld/fluxnet_tsne_strength_3_sub.csv}; 
    \end{axis} 

\node[text width=1cm, inner sep =2em] (nleftup) at (0.3,2.75) { \renewcommand{\arraystretch}{.7}   
\begin{tabular}{l}   \tiny{medium$\,P$}\\ \tiny{medium$\,R_g$}\end{tabular}};      
\node[ ] (nrightup) at (2.55,2.75) { \renewcommand{\arraystretch}{.7}   
\begin{tabular}{l}   \tiny{high $P$}\\ \tiny{high $R_g$}
\end{tabular}};  
\node[ ] (nleftdn) at (0.55,0.3){ \renewcommand{\arraystretch}{.7}   
\begin{tabular}{l}   \tiny{low $P$}\\ \tiny{medium$\,R_g$}
\end{tabular}};  
\node[ ] (nrightdn) at (2.55,0.3){ \renewcommand{\arraystretch}{.7}   
\begin{tabular}{l}   \tiny{low $P$}\\ \tiny{high $R_g$}
\end{tabular}};  
\end{tikzpicture}
        \caption{\label{fig:flux_b}
        $t$-SNE Embedding.} 
\end{subfigure} 
\caption{ \label{fig:flux} \emph{Biosphere-Atmosphere Interactions.} For the FLUXNET data from 64 locations $\D$ in multiple years $\T$, we show the DAG $\Gg$ (a) that \ourmethod discovers.  We visualize the causal strengths of the seven edges in $\Gg$ in two dimensions using $t$-SNE (b), where each sample $(x_1, x_2)$ corresponds to a fixed location $d_i$, month $m_j$, and year $y_k$. We recover regions  that correspond to distinct underlying meteorological conditions (precipitation $P$, global radiation $R_g$). 
}
\end{figure}
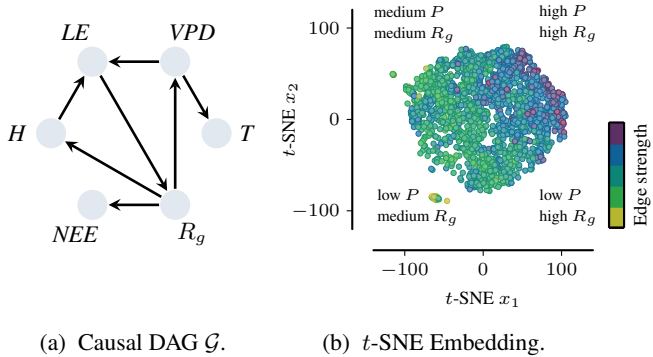

\subsection{Case Study: Reconstructing Gradients of Biosphere-Atmosphere Interactions}
In the FLUXNET datasets, we study interactions of meteorological and atmospheric variables, with a primary focus on air temperature (\emph{T}), global radiation ($R_g$), net ecosystem exchange (\emph{NEE}), sensible heat (\emph{H}), latent heat flux (\emph{LE}) and vapor pressure deficit (\emph{VPD}).

\paragraph{Causal Strengths across Datasets} 
We show the common graph jointly over all locations and months in Fig. \ref{fig:flux_a}.
As there is no known ground truth regarding the  network, 
we replicate the qualitative analysis in ~\citet{krich:21:flux} to see whether the causal strengths across locations and time
correspond to plausible interactions between atmospheric and meteorological variables. To this end, we consider the MDL edge strengths  $\delta E$ for the seven edges in $\G$ for each location $d \in \D$ and each month $m  \in \T_d$, and project them to two dimensions using the $t$-distributed stochastic neighbor embedding ($t$-SNE). The result is shown in Fig. \ref{fig:flux_b}, where each sample $(x_1, x_2)$ represents a location $d$ and month $m$. 

Perhaps surprisingly, the grouping by causal strength is not strongly tied to the actual geographic location and corresponding ecosystem, but rather correlates with water availability and global radiation. For example, we find strong edge connections in $\Gg$ (upper right) to be associated with high radiation and precipitation, even though these variables  are not directly used during causal discovery. 
As noted in prior work~\cite{kraemer:20:biospherestates,krich:21:flux}, these conditions create an optimal growing environment in the respective ecosystems and months. This opposes a low-energy output state often reached in the winter time, which we reconstruct with weak edge connectivity in $\Gg$ (lower left).

Grouping the biosphere-atmosphere interactions by causal strength therefore reveals different meteorological states that the ecosystems traverse over the year. In particular, matching previous insights~\cite{krich:21:flux} these findings support  the hypothesis that distinct ecosystems can traverse very similar meteorological states over time.

\tikzset{ 
    dag_node/.style={draw=none, thick, rounded corners, circle,
        minimum size=0.4cm, fill=gray!35}
}
\def\yshiftts{0}
\def\poscontextts{-1}
\def\poslabelts{\poscontextts}
\def\xshiftts{\poslabelts} 

\def\yrsRzeroDeHai{2000,2001,2002,2004,2005,2006,2007,2008} 
\def\yrsRoneDeHai{2003} 
\def\yrsRtwoDeHai{2009,2010,2011}
\def\yrsRzeroDeTha{2000,2001,2002,2004,2005,2007,2008} 
\def\yrsRoneDeTha{2003} 
\def\yrsRtwoDeTha{2006} 
\begin{figure}[t] 
    \begin{subfigure}[c]{0.22\textwidth} 
        \begin{tikzpicture}
        \begin{axis}[
            scale only axis,
            width=4cm,
            height=2.5cm,
            axis x line=none,  
            axis y line=none,  
            xtick=\empty,
            ytick=\empty, legend image post style={xscale=.4, yscale=1}, 
            xmin=0, legend style={font=\tiny},
            legend entries={$R_1$  (other),,,,,,,,$R_2$ (2003)}
        ]  
            \node[draw=none] at (axis cs:\poslabelts,0) {\emph{P}}; 
    \foreach \yr in \yrsRzeroDeHai{ 
        \addplot[slate200, mark=none,  
        ] table [x=t, y=P_F] {expres/flux_texdata/DE-Hai_\yr};  
    }
    \foreach \yr in \yrsRoneDeHai{ 
        \addplot[\colorcontextone, mark=none,  line width=1.5pt,
        ] table [x=t, y=P_F] {expres/flux_texdata/DE-Hai_\yr}; 
    }   
        \def\ytimeline{-7.1*\yshiftts}
       \addplot[draw=none] coordinates{(\xshiftts,\ytimeline) (\xshiftts+12,\ytimeline-1)};
        \draw[->] (axis cs:\xshiftts,\ytimeline) -- (axis cs:\xshiftts+12,\ytimeline);
    \node[draw=none] at (axis cs:\xshiftts+12.5,\ytimeline) {$t$};
        \end{axis}
        \end{tikzpicture}
        \caption{Regimes $\R$ in $d_1:$ De-Hai.}
        \end{subfigure}
          \begin{subfigure}[c]{0.22\textwidth} 
        \begin{tikzpicture}
        \begin{axis}[
            scale only axis,
            width=4cm,
            height=2.5cm,
            axis x line=none,  
            axis y line=none,  
            xtick=\empty,
            ytick=\empty, legend image post style={xscale=.4, yscale=1}, 
            xmin=0, legend style={font=\tiny},
            legend entries={$R_1$  (other),,,,,,,$R_2$ (2003),$R_3$ (2006)}
        ]  
            \node[draw=none] at (axis cs:\poslabelts,0) {\emph{P}}; 
    \foreach \yr in \yrsRzeroDeTha{ 
        \addplot[slate200, mark=none,  
        ] table [x=t, y=P_F] {expres/flux_texdata/DE-Tha_\yr};  
    }
    \foreach \yr in \yrsRoneDeTha{ 
        \addplot[\colorregimeone, mark=none,  line width=1.5pt,
        ] table [x=t, y=P_F] {expres/flux_texdata/DE-Tha_\yr}; 
    }  
    \foreach \yr in \yrsRtwoDeTha{ 
        \addplot[\colorregimetwo, mark=none,  line width=1.5pt,
        ] table [x=t, y=P_F] {expres/flux_texdata/DE-Tha_\yr}; 
    }   
        \def\ytimeline{-7.1*\yshiftts} 
        \draw[->] (axis cs:\xshiftts,\ytimeline) -- (axis cs:\xshiftts+12,\ytimeline);
        \addplot[draw=none] coordinates{(\xshiftts,\ytimeline) (\xshiftts+12,\ytimeline-1.2)};
        \node[draw=none] at (axis cs:\xshiftts+12.5,\ytimeline) {$t$};
        \end{axis}
        \end{tikzpicture}
        \caption{Regimes $\R$ in $d_2:$ De-Tha.}
        \end{subfigure}
    \caption{\emph{Regime partitioning reveals distinct trends.} The regime partitions that \ourmethod discovers reveal abnormal conditions, such as the effects of a European heatwave (2003) on the FLUXNET ecosysyems Hainich (De-Hai) and Tharandt (De-Tha) (2000-2009).}
    \label{fig:fluxregime}
\end{figure}
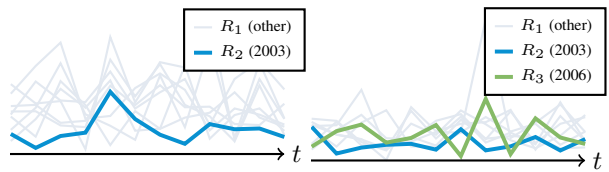

\paragraph{Distinct Trends over Time}  A limitation of the $t$-SNE visualization is that the resulting space is not metric and thus does not provide a direct way to perform comparisons~\cite{krich:21:flux}. With \ourmethodns, we can test for statistical differences in causal mechanism, and, for example, investigate how the causal interactions evolve over the years at a given location. To illustrate, we show in Fig.~\ref{fig:fluxregime} for the locations Hainich (DeHai) and Tharandth (DeThai) the regime partitions over multiple years (here 2000-2009). We discover a common causal mechanism in most years (greyed out) and distinct regimes in 2003, resp. 2006 (colored). Such changes likely result from abnormal meteorological conditions; in this case, a drought event in Europe in 2003, which is reflected in the monthly precipitation $P$ (Fig.~\ref{fig:fluxregime}).


\section{Conclusion}
While causal discovery often treats cause-effect relationships as fixed, most real-world processes change and evolve, often in multiple environments and over time. Thinking of such settings, we study causal discovery and changepoint detection for a collection of multivariate time series. We model causal mechanism shifts across datasets and time jointly using the notions of contexts and regimes. 
We propose  discovering temporal causal graphs using Minimum Description Length (MDL) encodings for Gaussian processes (GPs), where we detect regime changepoints from changes in residual distributions. Alongside this, we discover contexts and regimes with similar causal mechanisms using kernelized hypothesis testing. We confirmed that our strategy for discovering causal models, changepoints, and partitions, summarized in the \ourmethod algorithm, works well in our simulations. On real-world data, our method can be used to discover summary networks, varying edge strengths, and meaningful groups of time spans and locations with similar mechanisms. This supports a research direction where we base cluster analysis not only on the marginal distributions, but on causal interaction structures between variables of interest \cite{guenther:23:river}, which future continuations of this work could explore further. 


\section{Acknowledgments}
U.N. has received funding from the European Research Council (ERC) Starting Grant CausalEarth under the European Union’s Horizon 2020 research and innovation program (Grant Agreement No. 948112).

We thank the anonymous reviewers for their helpful comments.

\bibliography{bib/abbreviations,bib/bib-paper,bib/bib-jilles}
 
\end{document}